\pdfoutput=1

\documentclass[11pt]{article}

\usepackage[preprint]{acl}
\usepackage{bbding}
\usepackage{colortbl}
\usepackage{times}
\usepackage{latexsym}
\usepackage{bbm}
\usepackage[T1]{fontenc}

\usepackage[utf8]{inputenc}

\usepackage{microtype}

\usepackage{inconsolata}

\usepackage{graphicx}

\usepackage{amsmath}
\usepackage{pifont}
\usepackage{array}
\usepackage{booktabs}
\usepackage{enumitem}
\usepackage{amssymb}
\usepackage{multicol}
\newcommand{\vpara}[1]{\vspace{0.05in}\noindent \textbf{#1 }}
\newcommand{\bench}{VisualSimpleQA}
\newcommand{\sbench}{VisualSimpleQA\space}

\usepackage{threeparttable}
\usepackage{makecell}
\usepackage{graphicx}
\usepackage{multirow}
\usepackage{graphicx}
\usepackage{subfigure}
\usepackage{wrapfig}
\usepackage{graphicx}
\usepackage{caption}
\usepackage{amssymb}
\usepackage{hyperref}
\hypersetup{
    colorlinks = true,
    citecolor = {blue}, 
    urlcolor = {blue},
    linkcolor = {blue}  
}
\usepackage{tcolorbox}

\title{\bench: A Benchmark for Decoupled Evaluation of Large Vision-Language Models in Fact-Seeking Question Answering}

\author{
    Yanling Wang\textsuperscript{\rm 1},\ Yihan Zhao\textsuperscript{\rm 2},\ Xiaodong Chen\textsuperscript{\rm 2},\ Shasha Guo\textsuperscript{\rm 2}, \\ \ {\bf Lixin Liu}\textsuperscript{\rm 3}, \ {\bf Haoyang Li}\textsuperscript{\rm 2}, 
    \ {\bf Yong Xiao}\textsuperscript{\rm 1}, \ {\bf Jing Zhang}\textsuperscript{\rm 2}, \ {\bf Qi Li}\textsuperscript{\rm 1, 4}, \ {\bf Ke Xu}\textsuperscript{\rm 1, 4} \\
    \textsuperscript{\rm 1} Zhongguancun Laboratory
    \textsuperscript{\rm 2} Renmin University of China\\
     \textsuperscript{\rm 3} Tencent
    \textsuperscript{\rm 4} Tsinghua University\\
     \{wangyl, xiaoyong\}@ruc.edu.cn \quad kylenliu@tencent.com \\
    \{zhaoyihan, chenxiaodong, guoshashaxing, lihaoyang.cs, zhang-jing\}@ruc.edu.cn \\
    \{qli01, xuke\}@tsinghua.edu.cn
}

\begin{document}
\maketitle
\begin{abstract}
Large vision-language models (LVLMs) have demonstrated remarkable achievements, yet the generation of non-factual responses remains prevalent in fact-seeking question answering (QA). Current multimodal fact-seeking benchmarks primarily focus on comparing model outputs to ground truth answers, providing limited insights into the performance of modality-specific modules. To bridge this gap, we introduce \bench~\footnote{The dataset is available at \url{https://huggingface.co/datasets/WYLing/VisualSimpleQA}.}, a multimodal fact-seeking benchmark with two key features. First, it enables streamlined and decoupled evaluation of LVLMs in visual and linguistic modalities. Second, it incorporates well-defined difficulty criteria to guide human annotation and facilitates the extraction of a challenging subset, \bench-hard.
Experiments on 15 LVLMs show that even state-of-the-art models such as GPT-4o achieve merely 60\%+ correctness in multimodal fact-seeking QA on \sbench and 30\%+ on \bench-hard.
Furthermore, the decoupled evaluation across these models highlights substantial opportunities for improvement in both visual and linguistic modules.
\end{abstract}

\section{Introduction}
Alongside the progress in large language models (LLMs)~\cite{GPT-4, DeepSeek-V3}, large vision-language models (LVLMs)~\cite{gpt-4o,anthropic2024claude,deepmind2024gemini}
have also emerged as a cornerstone of artificial intelligence. 
Despite their success, generation of factually incorrect responses remains a key obstacle to their wider application.

To address this problem, a crucial step is to effectively evaluate the fact-seeking QA capabilities of these models. Current benchmarks~\cite{simpleqa,truthfulqa,naturalquestions, okvqa, MMStar, MMT-Bench, MFC-Bench} often use an end-to-end approach, directly comparing the outputs with the ground truth answers. However, due to the involvement of multiple modalities, incorrect outputs from an LVLM may stem from inaccurate visual recognition, insufficient relevant knowledge, or a combination of both. As a result, decoupled evaluation of an LVLM's fact-seeking QA capability is essential.

\begin{figure}[t]
	\centering
		\includegraphics[width=3in]{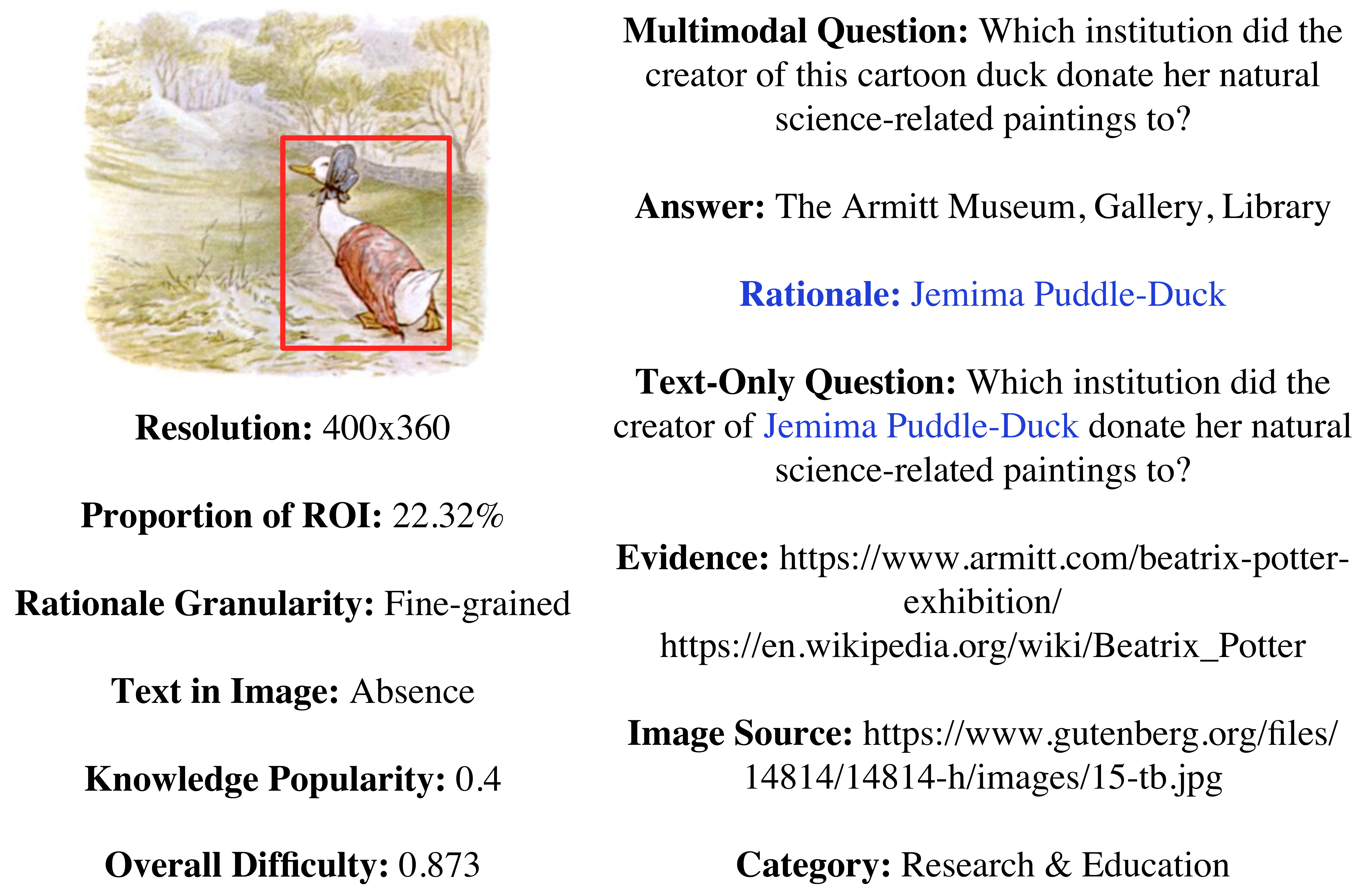}
      \caption{Illustration of an example in \bench. The red box highlights the region of interest (ROI). 
      Each sample has several attributes and tags, which allow us to measure its overall difficulty score based on our proposed difficulty criteria.}
  \label{fig:intro-examples}
\end{figure}

Furthermore, since large models advance rapidly, challenging samples are essential for evaluating cutting-edge models like GPT-4o~\cite{gpt-4o} and Claude-3.5-Sonnet~\cite{anthropic2024claude}. Recently, AI researchers and trainers at OpenAI developed a benchmark named SimpleQA~\cite{simpleqa} to evaluate the factuality of LLMs, on which both GPT-4o and Claude scored below 50\% correctness. 
However, there is a lack of well-defined difficulty criteria to guide the annotation process, particularly within the LVLM community.

To address these limitations, we introduce a multimodal fact-seeking QA benchmark named \sbench and have gathered a team of human annotators to create the samples. Figure~\ref{fig:intro-examples} presents an example. Specifically, \sbench exhibits the following properties:

\begin{itemize}
    \item \textbf{Decoupled evaluation.} 
    In addition to the multimodal question and the ground truth answer, \sbench includes another two annotations: the rationale for answering the multimodal question and a rewritten text-only question. The rationale provides a concise description of the region of interest (ROI), i.e., the information that needs to be extracted from the image for answering the question. With the rationale, we reformulate the multimodal question into a self-contained, text-only question that does not rely on any visual input. Based on the text-only questions, we evaluate the linguistic module's capability in fact-seeking QA. Then we evaluate the visual module according to the relative performance degradation when transitioning from text-only QA to multimodal QA.
    \item \textbf{Well-defined difficulty criteria.} 
     Compared to previous multimodal fact-seeking QA benchmarks~\cite{okvqa, A-OKVQA}, \sbench introduces well-defined difficulty criteria that quantify each sample's challenge from both visual and linguistic perspectives. 
     These criteria are associated with attributes and tags illustrated below the image in Figure~\ref{fig:intro-examples}.
     These criteria not only guide human annotation but also facilitate the extraction of \bench-hard, a set containing 129 particularly challenging samples. On \bench-hard, GPT-4o and Claude-3.5-Sonnet score only 30\%+ correctness. This challenging set is instructive for evaluating cutting-edge models.
    \item \textbf{High quality \& diversity.}
    All samples are created by humans with at least one year of experience working with large models. The ground truth answers are accurate and precise, supported by evidence from formal or authoritative sources. Additionally, we provide image sources, either as URLs or the names of public datasets.
    Furthermore, dedicated inspectors perform at least two quality checks on each sample. \sbench also covers a broad spectrum of topics, including research \& education, company \& brand, film \& television entertainment, politics, history, and more, ensuring its diversity.
    \item \textbf{Reduced bias on evaluation.}
    Several multimodal benchmarks~\cite{okvqa, A-OKVQA} are built based on public image datasets like COCO~\cite{COCO}. These publicly available datasets are likely to be used for LVLM training, which may result in bias on evaluation. To mitigate this issue, images for 200 samples (40\% of \bench) are newly collected from the Internet.
    
\end{itemize}

We evaluate fifteen leading closed-source and open-source LVLMs using \sbench and \bench-hard. The results reveal substantial opportunities for improving LVLMs for multimodal fact-seeking QA, due to their limitations in handling complex visual recognition tasks and/or the lack of long-tailed knowledge.

This paper makes the following contributions:

\begin{itemize}
    \item We introduce a benchmark named \bench, which enables a streamlined, decoupled evaluation of LVLMs in fact-seeking QA, simplifying the assessment of modality-specific modules in LVLMs.
    
    \item \sbench is paired with well-defined difficulty criteria for samples. Based on the criteria, we extract \bench-hard, a challenging sample set for evaluating the factuality of cutting-edge LVLMs.
    
    \item Decoupled evaluation of 15 frontier LVLMs highlights substantial room for improvement in both linguistic and visual modules for fact-seeking QA. State-of-the-art models such as GPT-4o achieve only 60\%+ correctness on multimodal questions in \sbench and 30\%+ on those in \bench-hard.
\end{itemize}

\section{Related Work}
\vpara{Decoupled Evaluation of LVLMs.}
The rapid advancement of LLMs has driven the development of LVLMs~\cite{gpt-4o,anthropic2024claude,deepmind2024gemini,Molmo-7B-O-0924,qwen2024qvq}.
Due to the multimodal nature of LVLMs, it is non-trivial to evaluate each modality-specific module. To address this, HallusionBench~\cite{Hallusionbench} was designed to attribute hallucinated outputs into language hallucinations, visual illusions, or a combination of both. However, the benchmark  pairs some images with edited versions for evaluation and relies on a complex decision tree to diagnose failure types, which may reduce its practicality. Another approach, Prism~\cite{Prism}, was proposed to decouple and assess the capabilities of LVLMs. Yet, Prism requires an LLM as a reference during evaluation, and as the authors analyzed, this approach may be less effective for state-of-the-art LVLMs. In contrast, \sbench provides a more streamlined approach for decoupled evaluation.

\begin{table}
    \small
   \centering
   \renewcommand\arraystretch{0.9}
   \begin{threeparttable}
    \newcolumntype{?}{!{\vrule width 0.5pt}}
    \setlength{\tabcolsep}{1.8mm}
    \begin{tabular}{l?ccc}
        \toprule
        & \textbf{\makecell{Factuality-\\aware}} & \textbf{\makecell{Decoupled \\Evaluation}}  & \textbf{\makecell{Difficulty \\Criteria}} \\
        \midrule
        OK-VQA& All & No & No \\
        A-OKVQA & All & No & No \\
        MMBench & Partial & No & No \\
        MMStar & Partial & No & No \\
        MMT-Bench & Partial & No & No \\
        MFC-Bench & Partial & No & No \\
        Prism & -- & Yes & No \\
         HallusionBench & Partial & Yes & No\\
        \midrule
        \bench & All & Yes & Yes \\
        \bottomrule
    \end{tabular}
    \small
    \begin{tablenotes}
       \item Note:
       Prism is an evaluation framework that does not include newly constructed datasets.
    \end{tablenotes}
    \end{threeparttable}
    \caption{Comparison of properties between \sbench and relevant multimodal benchmarks.}
    \label{tab: Comparison of properties between different benchmarks}
\end{table}

\vpara{Factuality Evaluation of LVLMs.}
In terms of LLMs, several benchmarks have been developed for factuality evaluation, including SimpleQA~\cite{simpleqa}, TruthfulQA~\cite{truthfulqa}, Natural Questions~\cite{naturalquestions}, etc. In contrast, the most commonly used benchmarks for factuality evaluation of LVLMs are still the classic OK-VQA~\cite{okvqa} and A-OKVQA~\cite{A-OKVQA}. Recently, new multimodal benchmarks, such as MMBench~\cite{MMBench}, MMStar~\cite{MMStar}, MMT-Bench~\cite{MMT-Bench}, and MFC-Bench~\cite{MFC-Bench}, have been introduced. 
While these benchmarks involve factual knowledge, they primarily focus on a broader range of visual understanding tasks, rather than offering a fine-grained evaluation of the model's factuality.
Compared to these benchmarks, \sbench supports decoupled evaluation, has well-defined difficulty criteria for samples, and provides evidences of answers to enhance the data quality.

We summarize the comparison of properties between \sbench and the aforementioned typical multimodal benchmarks in Table~\ref{tab: Comparison of properties between different benchmarks}.

\section{Key Features of \bench}\label{sec: decoupled evaluation}
The primary motivation behind \sbench is to evaluate modality-specific modules in an LVLM and to propose clear difficulty criteria that support the annotation of instructive and challenging samples. In this section, we will outline the process of conducting decoupled evaluations and describe how we establish the difficulty criteria.

\begin{figure}
	\centering
		\includegraphics[width=3in]{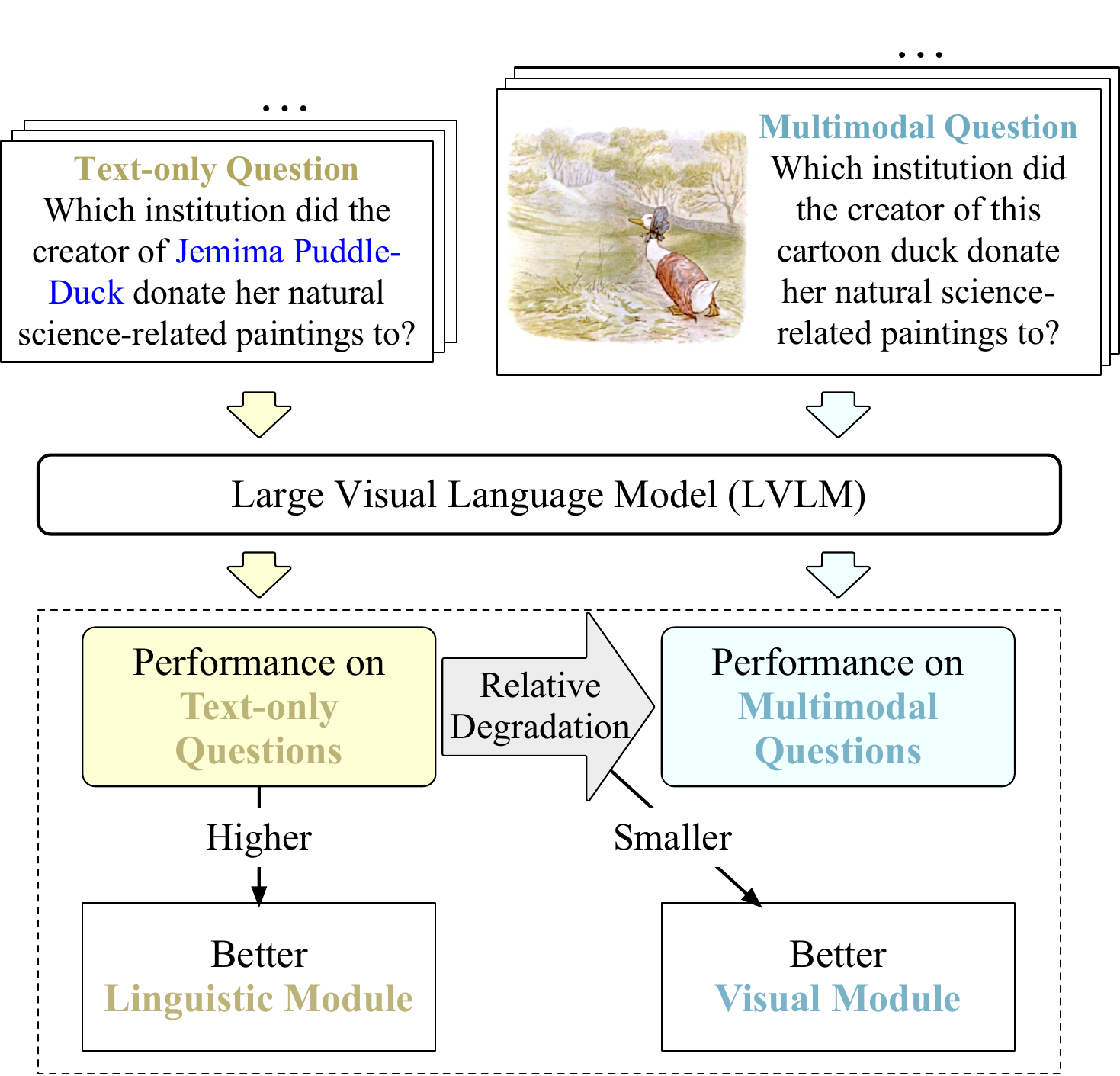}
      \caption{Decoupled evaluation process.
      }
  \label{fig:decoupled evaluation process}
\end{figure}

\subsection{Decoupled Evaluation}\label{sec: decoupled evaluation}
Decoupled evaluation, which aims to evaluate the modality-specific modules, is a key motivation behind \bench. 
The design of \sbench (see Figure~\ref{fig:intro-examples}) aligns with this goal. Each sample contains a multimodal fact-seeking question, a self-contained text-only question, and the ground truth answer for evaluation. Figure~\ref{fig:decoupled evaluation process} illustrates the decoupled evaluation process.

\vpara{Evaluation of Linguistic Module.} 
The performance of the LLM underlying an LVLM largely influences the capabilities of the LVLM. Stronger performance on text-only questions indicates that the model has a more effective linguistic module for fact-seeking QA. Conversely, if the model shows poor performance on the text-only questions, developers should enhance the knowledge in the linguistic module.

\vpara{Evaluation of Visual Module.} 
We calculate the performance degradation of each LVLM when transitioning from text-only QA (TQA) to multimodal QA (MQA). Since the rationale (i.e., the necessary information that needs to be extracted from an image) contained in the text-only question can be regarded as the ground truth for visual recognition, a smaller degradation indicates better performance of the visual module.
To facilitate comparisons across different models, we introduce a metric called relative degradation (RD),
\begin{equation}
    RD = \frac{score_{\text{TQA}} - score_{\text{MQA}}}{score_{\text{TQA}}} \label{eq: relative_degradation}
\end{equation}
where $score_{\text{TQA}}$ and $score_{\text{MQA}}$ represent the performance scores in text-only QA and multimodal QA, respectively. This metric quantifies the degree of performance degradation when an LVLM relies on its own visual recognition capabilities.

\subsection{Difficulty Criteria and Measurement}\label{sec: difficulty criteria and assessment}
In addition to supporting decoupled evaluation, \sbench aims to introduce well-defined criteria for measuring samples' difficulty levels.

\subsubsection{Difficulty Criteria}
As an LVLM involves both visual and linguistic modules, we establish a set of difficulty criteria from two perspectives—visual recognition and knowledge identification.

\vpara{Resolution.}
Resolution is the number of pixels in an image.
Higher-resolution images contain more pixels, making it easier for models to precisely identify visual features such as edge and texture.

\vpara{Proportion of ROI.} 
To correctly answer a multimodal question, the model requires precise visual grounding, i.e., the ability to identify the ROI. However, LVLMs may struggle to do this when the ROI~\footnote{This work focuses on multimodal questions involving a single ROI, while questions requiring multiple ROIs are left for future work.} is small or unclear~\cite{V-star-Saining}. 
Therefore, we use the proportion of ROI as a metric, calculated as the resolution of the ROI divided by that of the original image. 
A larger proportion of ROI typically facilitates visual recognition.

\vpara{Rationale Granularity.} 
We define the essential information or visual target extracted from the ROI as the rationale.
The rationale for answering a question can be either coarse-grained or fine-grained. For example, one question may ask to identify the animal species ``panda'' (coarse-grained), while another might ask for a specific cartoon character, such as ``Pikachu'' (fine-grained).
Intuitively, coarse-grained rationales are easier to identify.

\vpara{Presence or Absence of Text in Image.}
An image may contain text that help identify the rationale. Since latest LVLMs typically perform well on the optical character recognition (OCR) task, the recognized text from the image can naturally ease the rationale identification.
For example, the ROI in an image shows a zoo with the name ``Singapore Zoo'' written on its entrance gate. If an LVLM effectively performs OCR, the recognized text can serve as a cue for identifying the rationale ``Singapore Zoo''.
Therefore, we consider samples that can benefit from OCR to be easier.

\vpara{Knowledge Popularity.}
The training corpora for LLMs, such as CommonCrawl~\cite{commoncrawl}, The Stack~\cite{kocetkov2022stack}, and The Pile~\cite{gao2020pile}, are sourced from the Internet and likely contain a substantial amount of popular knowledge, which models can learn more effectively.
Conversely, long-tailed knowledge is typically more difficult for language models to grasp.
In this work, we prompt GPT-4o to assess the knowledge popularity for each sample, as it is a leading model well trained on a wide range of corpora, making it ideal for evaluating the popularity of the knowledge required to answer a question.
The prompt is provided in Appendix~\ref{appendix: KNOWLEDGE_POPULARITY_PROMPT}.

\subsubsection{Difficulty Measurement} 
According to the above criteria, the overall difficulty of a sample is defined as,
\begin{equation}
\small
    D = \text{Avg}\left(1 - R_{\text{norm}}, 1 - PR, \mathbbm{1}_{[\text{fine}]},  \mathbbm{1}_{[\text{w/o}\, \text{TI}]}, 1 - KP\right)
\label{eq: overall_difficulty}
\end{equation}
where Avg() denotes the average function, $R_{\text{norm}}$ denotes the Min-Max normalized resolution, $PR$ denotes the proportion of the ROI, and $KP$ denotes the knowledge popularity. The function $\mathbbm{1}$ is the indicator function. $\mathbbm{1}_{[\text{fine}]}$ takes a value of 1 if the rationale of a sample is fine-grained, and 0 otherwise. Similarly, $\mathbbm{1}_{[\text{w/o}\, \text{TI}]}$ takes a value of 1 if the image does not contain text that help identify the rationale, and 0 otherwise.
$D$ takes a value between 0 and 1.

\section{Annotation and Verification}
\begin{figure*}
	\centering
		\includegraphics[width=6in]{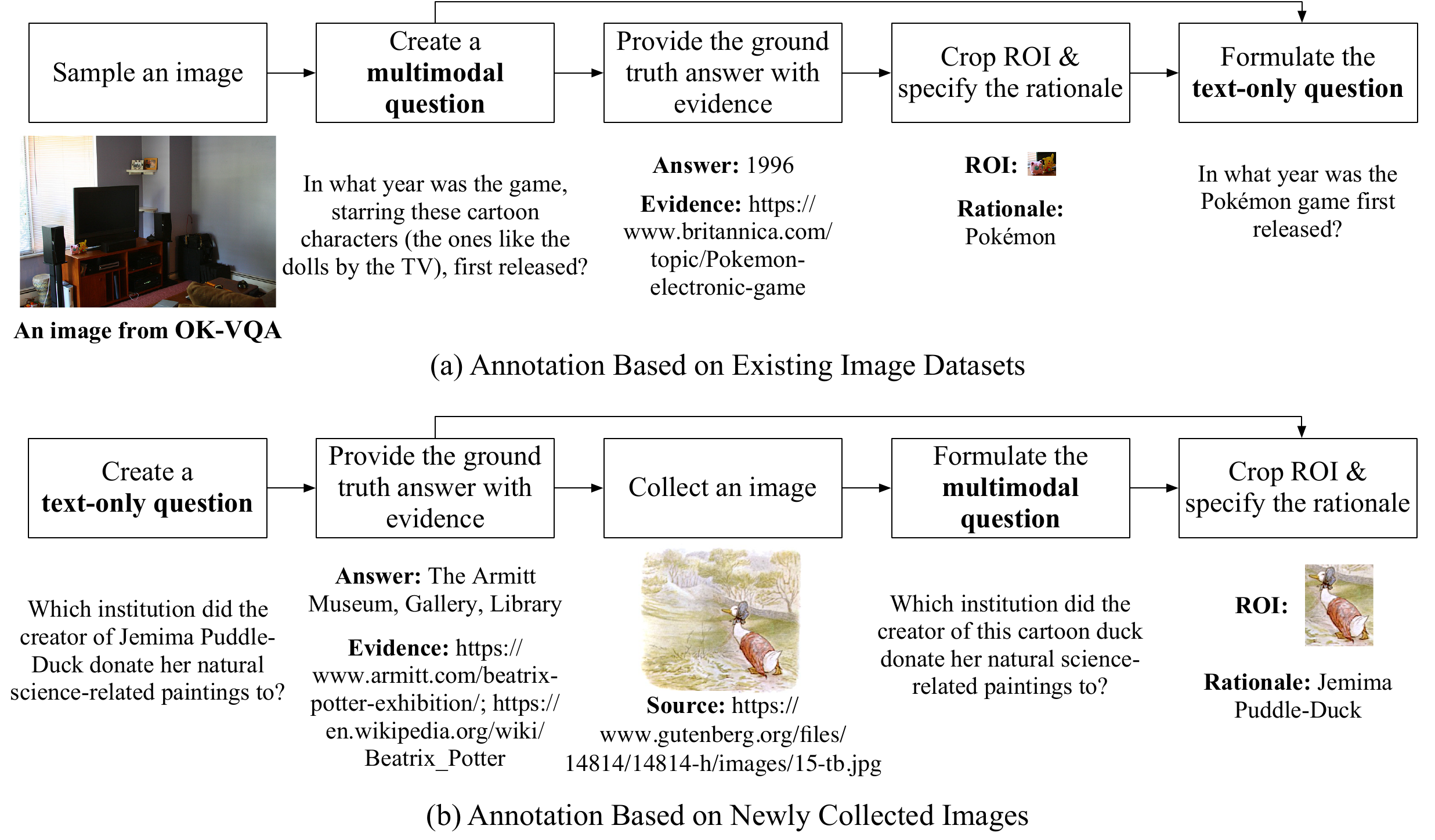}

      \caption{Flowchart of the annotation process. Evidence is used to guarantee the correctness of the answer, while ROI is annotated to calculate the difficulty of each sample.}
  \label{fig:flowchart}
\end{figure*}

\sbench is meticulously annotated and verified by six human annotators, each with at least one year of experience working with large models.

\subsection{Annotation Process}\label{subsec: annotatopn process}
Based on the image sources, we designed two methods to facilitate the annotation process. Figure~\ref{fig:flowchart} illustrates the flowchart of the annotation process.

\vpara{Annotation Based on Existing Image Datasets.}
An image from OK-VQA~\cite{okvqa} or V$^{*}$~\cite{V-star-Saining} is randomly sampled and presented to the annotator. According to the image and the difficulty criteria, the annotator creates a new fact-seeking \textit{multimodal question}.~\footnote{If the annotator cannot create a satisfactory question, they may opt to randomly sample another image.} In addition to the multimodal question, the annotator is required to provide the \textit{ground truth answer} with \textit{evidence}, crop the \textit{ROI} from the original image, specify the essential information to be extracted from the ROI (i.e., the \textit{rationale}), and reformulate the multimodal question into a self-contained, \textit{text-only question} that does not rely on any visual input.

\vpara{Annotation Based on Newly Collected Images.}
As images from existing benchmarks might have been used for model training, evaluation results based solely on these images could introduce bias. To mitigate this, we encourage annotators to collect new images from the Internet. Since it is not easy to determine which image to collect, we design another annotation method. Specifically, the annotator begins by creating a \textit{text-only question} along with its \textit{ground truth answer} and the \textit{evidence}. Next, the annotator searches for a suitable \textit{image}, reformulate the text-only question into a \textit{multimodal question}, crop the \textit{ROI} from the image, and specify the \textit{rationale}.

\vpara{Annotation Notes.}
Annotators should consider the following points:
\textbf{(1) Difficulty control.} We do not require every sample to be challenging for the advanced GPT-4o, as we hope that the difficulty levels of the samples are differentiated.
\textbf{(2) Knowledge cutoff.} The questions are set to be answerable by December 31, 2023, taking into account the knowledge cutoffs of various models.
\textbf{(3) Indisputable, short-form, and static answers.} Following SimpleQA~\cite{simpleqa}, annotators are asked to create questions that have indisputable and short-form answers, which promote more objective and accurate assessments, even under the LLM-based automatic evaluation framework~\cite{simpleqa, truthfulqa}. For example, instead of asking, ``When was version 7.22 of this software for Mac released?'', the question specifies, ``On what day, month, and year was version 7.22 of this software for Mac released?''
Additionally, the answer should remain consistent over time.
\textbf{(4) Traceability.} Each sample is annotated based on at least one webpage that contains the fact required by answering the question. The webpages must be formal or authoritative. Furthermore, the source of each image is provided. If the image is from existing benchmarks, the source is the name of the benchmark; if the image is newly collected by the annotator, the source is the URL of the image.
\textbf{(5) Diversity.} Annotators are requested to create samples involving diverse topics.

\subsection{Verification and Tagging}
\vpara{Verification.}
We engaged two AI researchers as checkers to examine all samples. Their task is to verify whether each sample met the following requirements: (1) adherence to the annotation notes mentioned in Section~\ref{subsec: annotatopn process}, (2) correct grammar, (3) absence of unsafe topics, and (4) accurate ground truth answer (the annotated evidence is a useful reference). Each checker reviewed all the samples, and if a sample was deemed improper by a checker, the two checkers collaborated to improve it.

\vpara{Tagging.}
The tags of a sample include rationale granularity, presence or absence of text in image, and topic category.
An annotator tags the topic category for each sample.
Two annotators tag the rationale granularity, with each annotator responsible for half of the samples.
Similarly, another two annotators tag the presence or absence of text in image, with each responsible for half of the samples. 
To minimize ambiguity in annotators' understanding of rationale granularity, we explicitly define questions about individuals, characters, organizations, locations, scientific concepts, or artworks as those requiring fine-grained rationales.

\section{Statistics of \bench}
\begin{figure}
	\centering
		\includegraphics[width=2.8in]{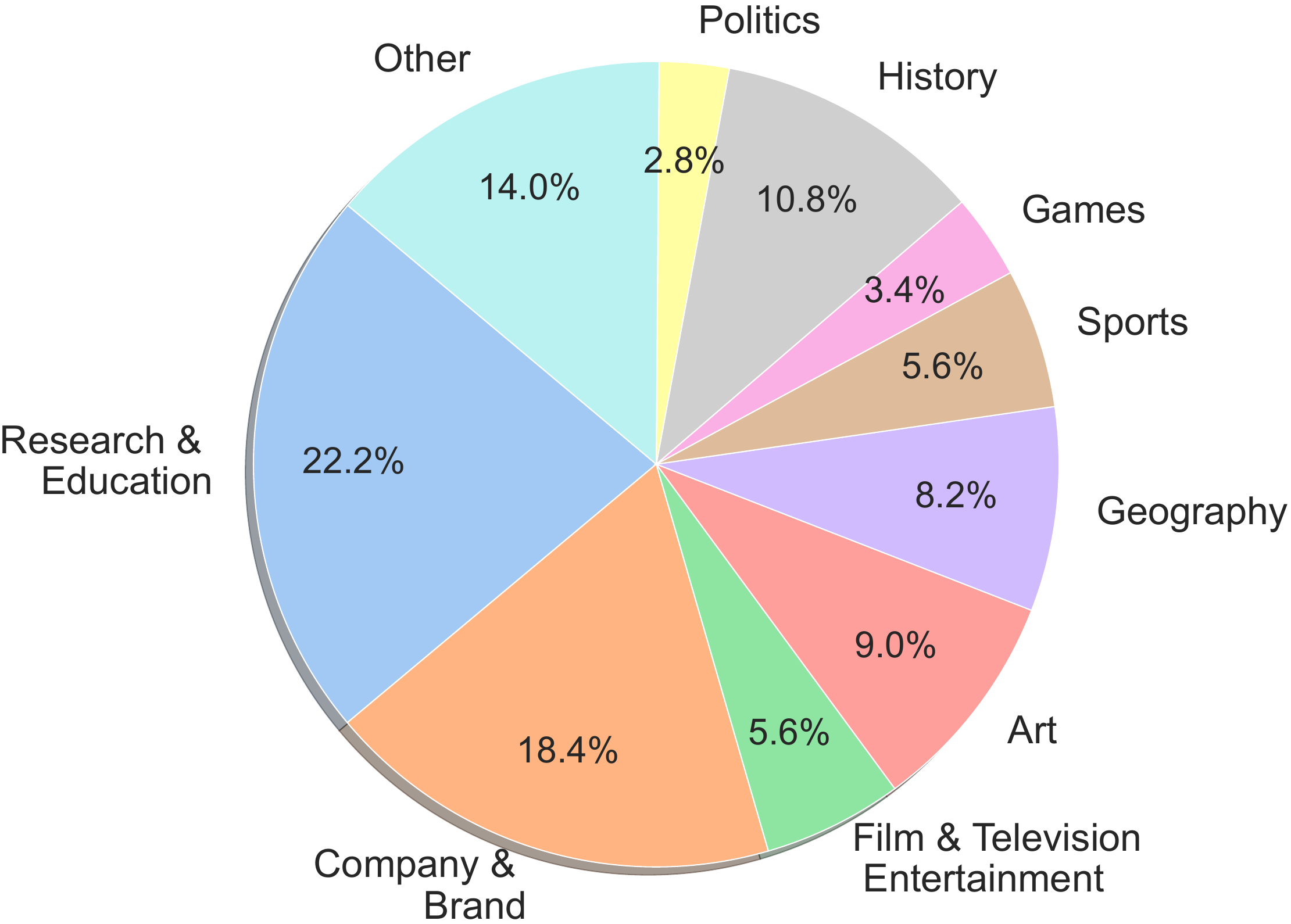}
      \caption{Distribution of topics in \bench.}
  \label{fig:diversity distribution}
\end{figure}

\sbench comprises 500 annotated samples, including 300 images from existing image datasets and 200 newly collected images from the Internet. In this section, we provide an overview of the statistical properties of \bench, focusing on its topic diversity and data difficulty.

\subsection{Topic Diversity}
\sbench covers a wide range of question topics, including the following topic categories: research \& education, company \& brand, film \& television entertainment, art, geography, sports, games, history, and politics. Samples that do not fit into these categories are classified as ``other.'' Figure~\ref{fig:diversity distribution} illustrates the distribution of topics in \bench.
We display examples of different categories in Appendix~\ref{sec:appendix_examples}.

\begin{figure}
	\centering
		\includegraphics[width=1.49in]{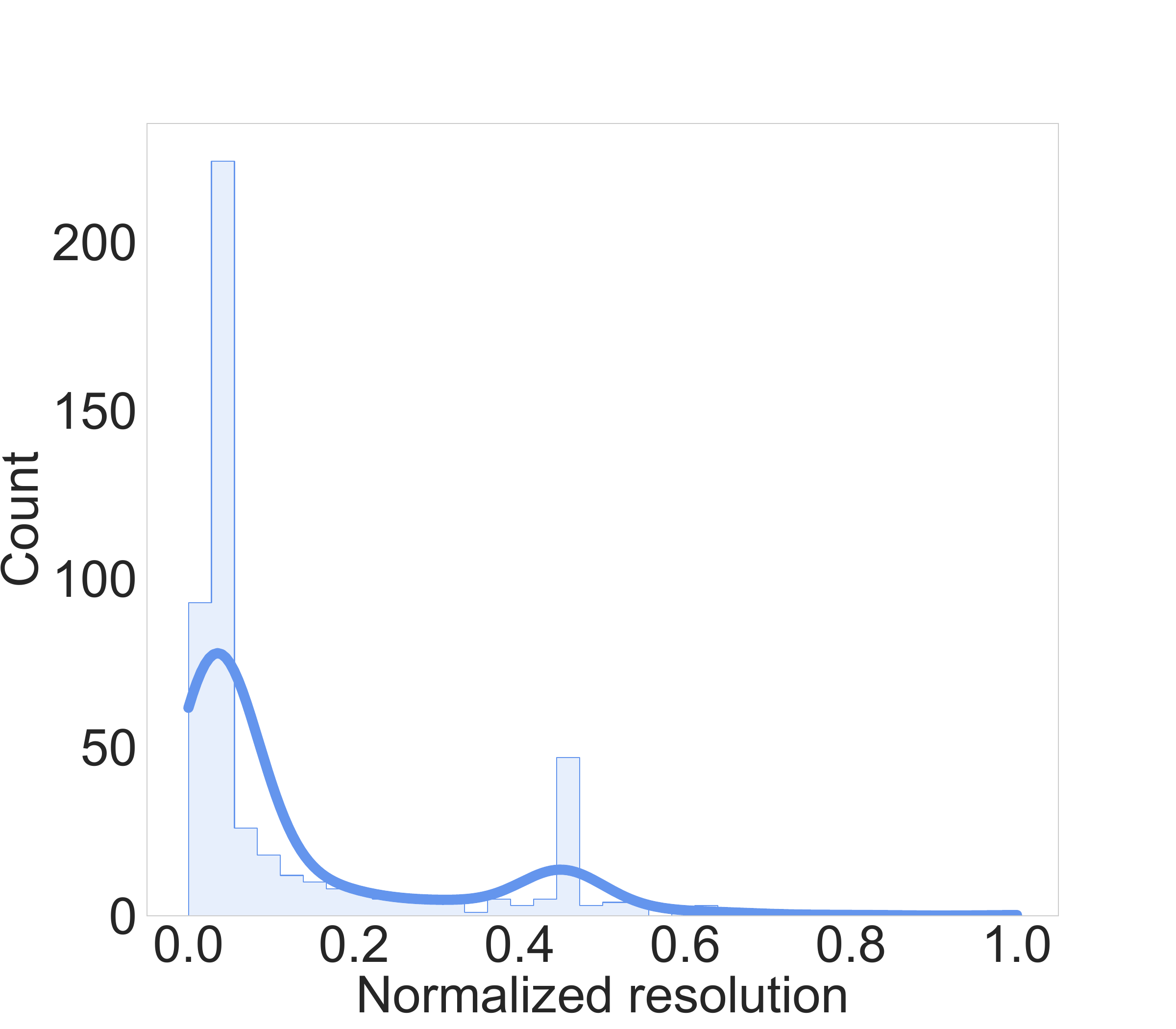}
        \includegraphics[width=1.49in]{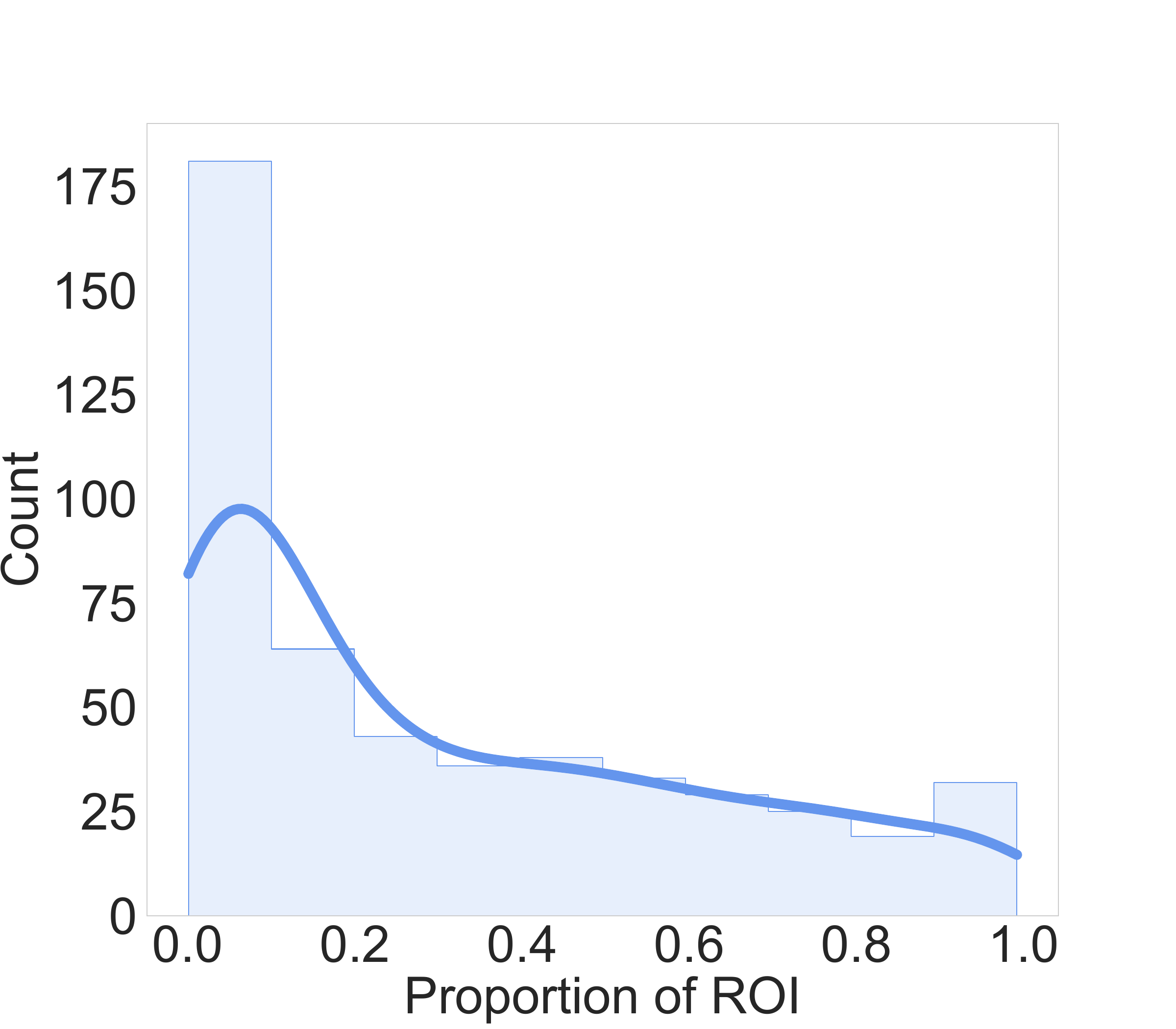}
        \includegraphics[width=1.49in]{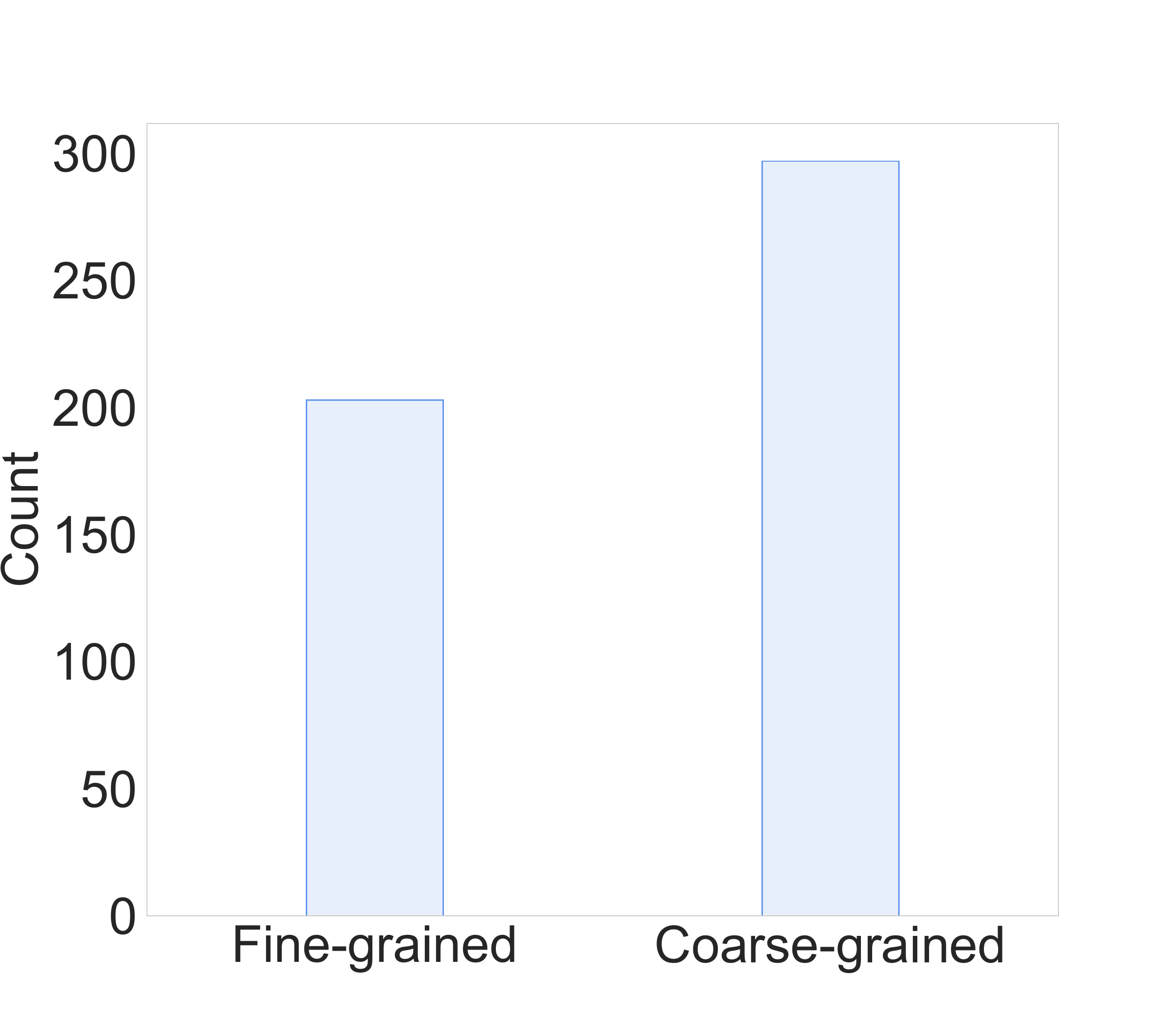}
        \includegraphics[width=1.49in]{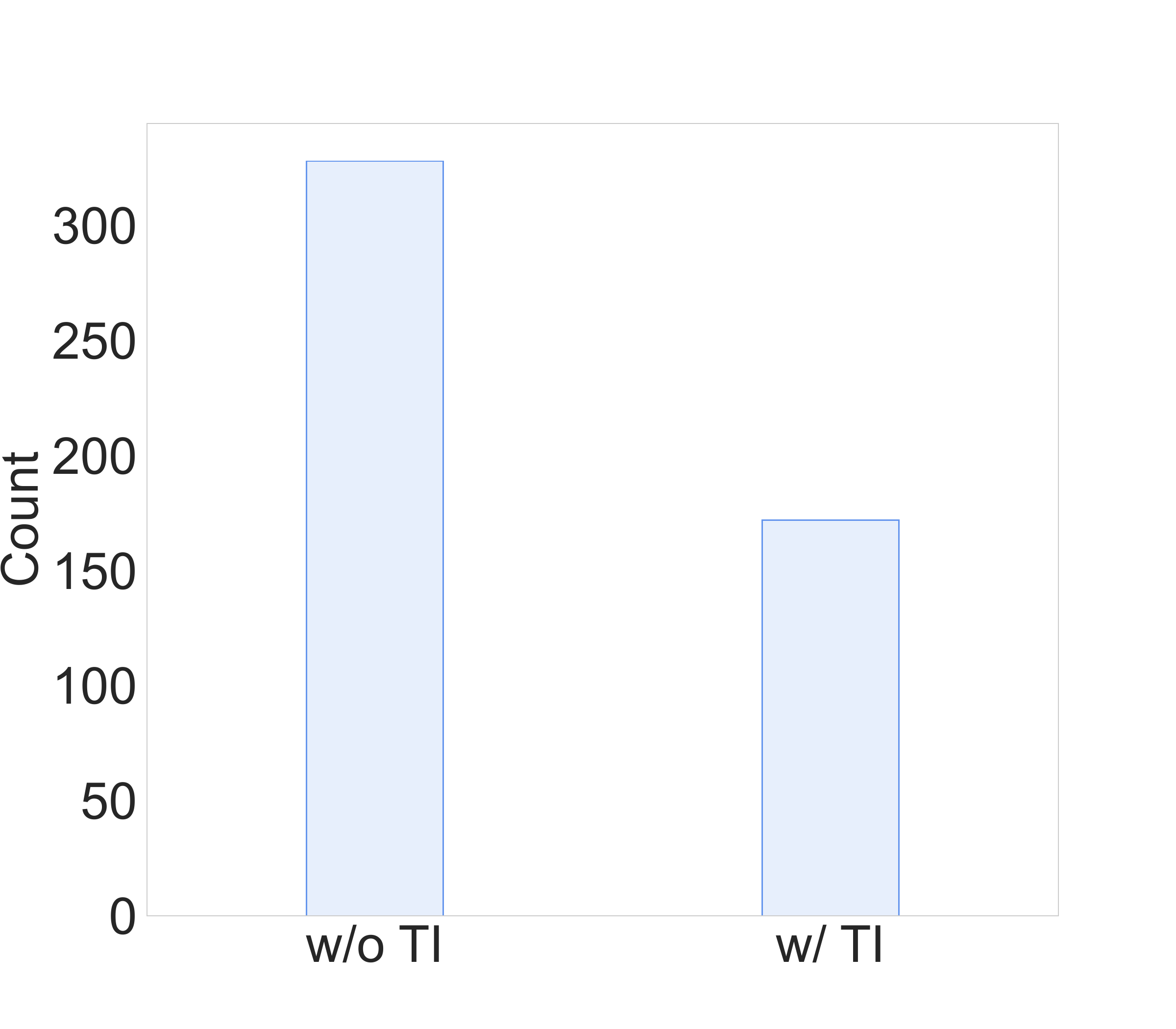}
      \caption{Distributions of factors that influence the difficulty of visual recognition. TI denotes Text in Image.}
  \label{fig:metric values distribution in visual recognition}
\end{figure}

\begin{figure}
	\centering
		\includegraphics[width=1.49in]{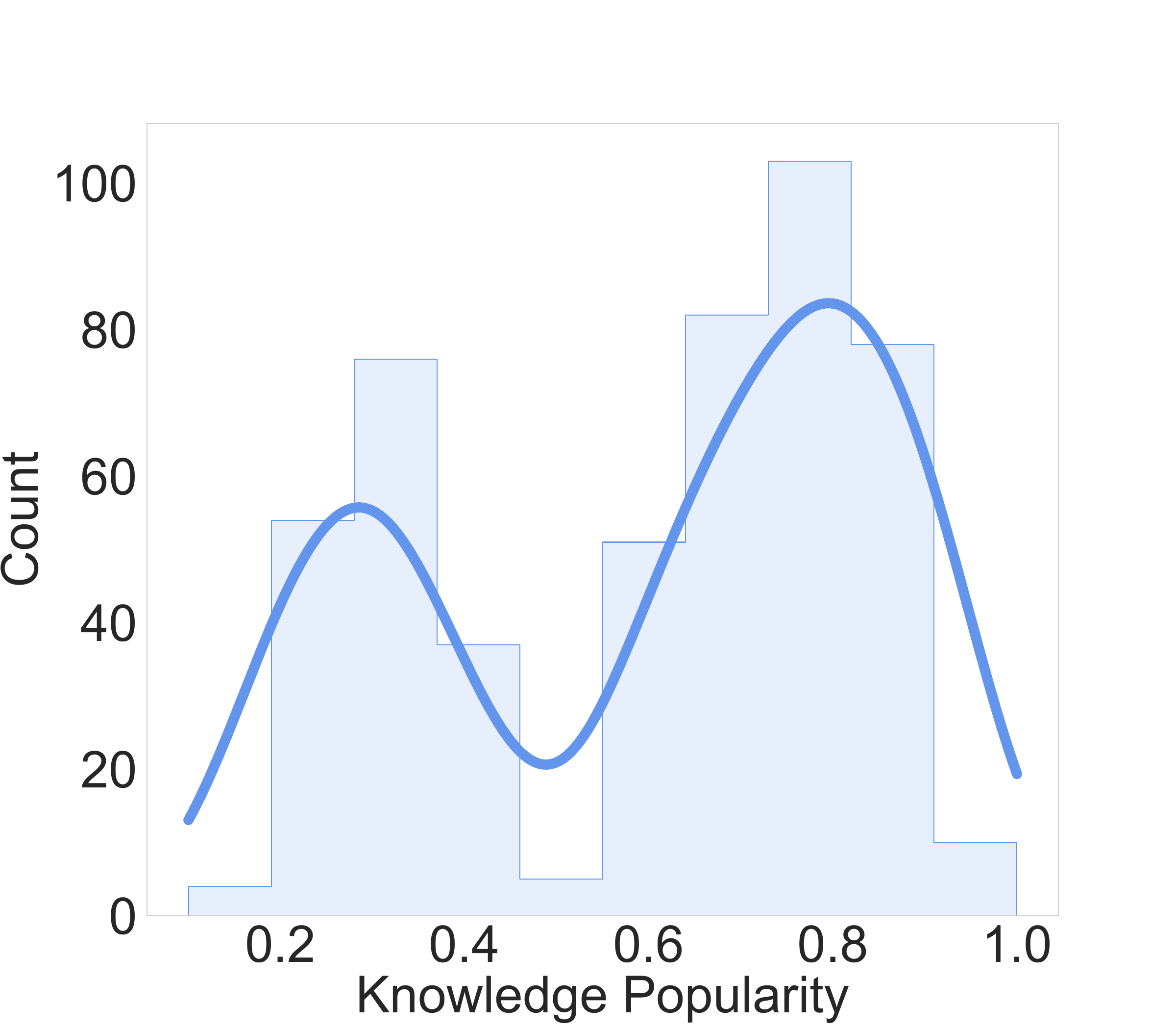}
        \includegraphics[width=1.455in]{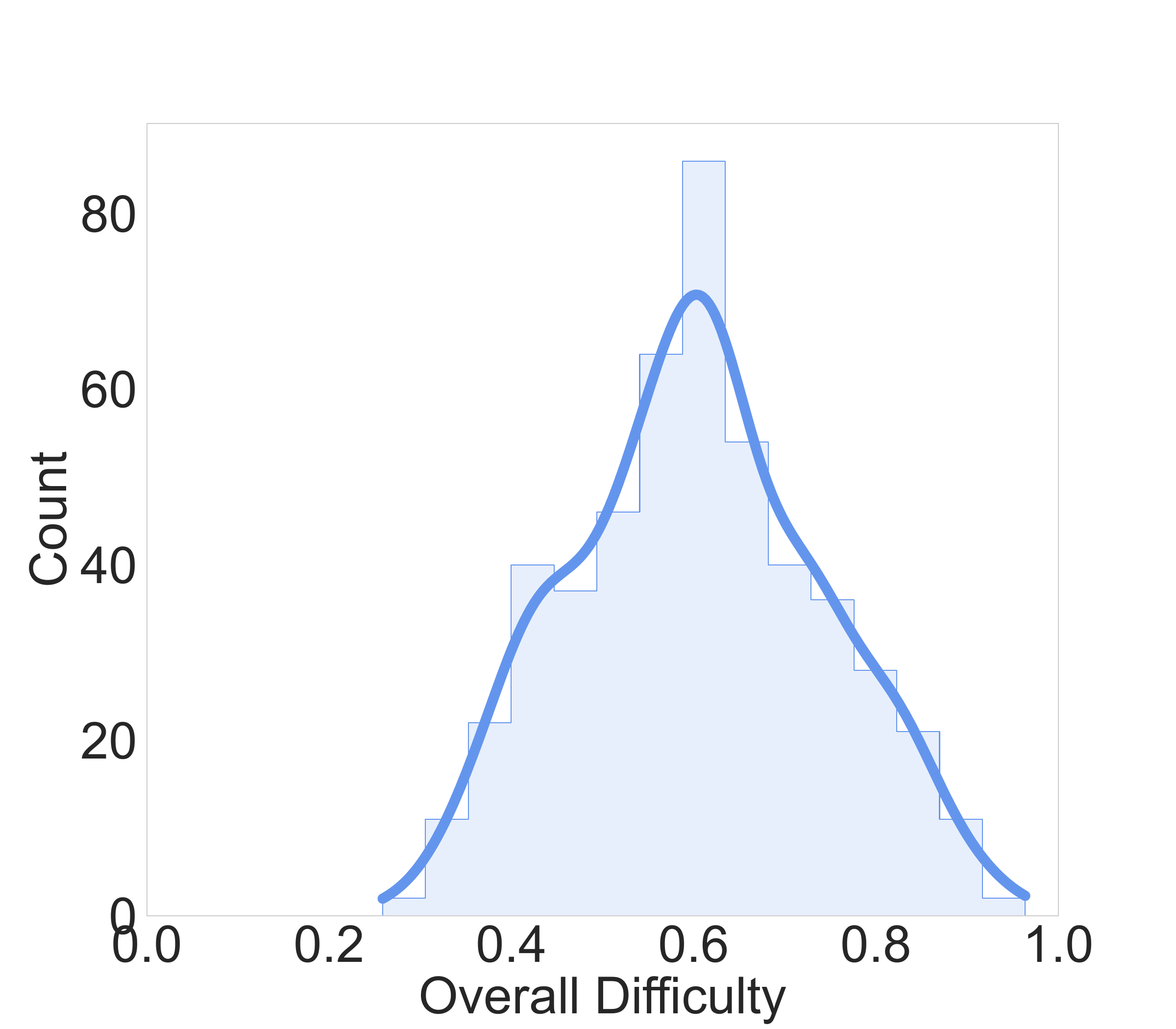}
      \caption{Distributions of knowledge popularity and overall difficulty.}
  \label{fig:knowledge popularity distribution}
\end{figure}

\subsection{Data Difficulty}
\vpara{Difficulty from a Visual Perspective.}
As shown in Figure~\ref{fig:metric values distribution in visual recognition}, most samples in \sbench are not of high resolution. Additionally, in most samples, the ROI, which reveals the rationale, constitutes a small area of the entire image. In terms of rationale granularity, 40.6\% of the samples feature fine-grained rationales. Furthermore, most images do not contain text, indicating the absence of direct cues for identifying the rationales.
These characteristics highlight that \sbench involves challenging visual recognition tasks.

\vpara{Difficulty from a Linguistic Perspective.}
As shown on the left side of Figure~\ref{fig:knowledge popularity distribution}, 35.2\% of the samples have a knowledge popularity score of 0.5 or lower, suggesting that at least 35.2\% of the knowledge in \sbench is considered not popular, even for advanced models like GPT-4o.

\vpara{Distribution of Overall Difficulty.}
The right side of Figure~\ref{fig:knowledge popularity distribution} shows that the majority of samples have overall difficulty scores above 0.5, indicating the challenging nature of the benchmark. Besides, the benchmark encompasses a range of difficulty levels, enabling a more instructive evaluation.
We select 129 samples with overall difficulty scores of 0.7 or higher to form the \bench-hard subset, which focuses on more challenging cases.

\section{Experiments}\label{sec: exp}
\subsection{Settings}
\vpara{Models.} We conduct evaluations on 15 frontier LVLMs, including open-source and closed-source models. The closed-source models include GPT-4o~\cite{gpt-4o}, Claude-3.5-Sonnet-20241022~\cite{anthropic2024claude}, and Gemini-2.0-flash-exp~\cite{deepmind2024gemini}. The open-source models include Molmo-7B-O-0924~\cite{Molmo-7B-O-0924}, LLaVA-OneVision--qwen2-7b-ov~\cite{LLaVA-OneVision}, Qwen2-VL-7B-Instruct~\cite{Qwen2-VL}, Qwen2.5-VL-7B-Instruct~\cite{Qwen2.5}, InternVL2.5-8B~\cite{InternVL2.5}, Llama-3.2-11B-Vision-Instruct~\cite{llama3.2}, Pixtral-12B-2409~\cite{Pixtral12B}, as well as larger models: Molmo-72B-0924~\cite{Molmo-7B-O-0924}, Qwen2-VL-72B-Instruct~\cite{Qwen2-VL}, Qwen2.5-VL-72B-Instruct~\cite{Qwen2.5}, QVQ-72B-Preview~\cite{qwen2024qvq}, and InternVL2.5-78B~\cite{InternVL2.5}.
Modules in these LVLMs
are shown in Appendix~\ref{appendix:different modules in LVLMs}.

\vpara{Grading and Metrics.}
Given the excellent performance and low API cost, we use DeepSeek-V3 ~\cite{DeepSeek-V3} as a classifier to classify the LVLM-generated responses into: ``correct'', ``incorrect'', or ``refusal''.~\footnote{We randomly selected 100 evaluations across different models and assessed the effectiveness of the automatic grading by human verification. 94\% of them were graded accurately.} The prompt is provided in Appendix~\ref{appendix: Prompts for Automatic Evaluation by DeepSeek-V3}.
Based on this, we can calculate the percentage of each grade. Notably, ``refusal'' means the LVLM acknowledges its inability to answer a question and thus mitigate the risk of hallucination~\cite{zhang2023siren, hallucinationSurveyHuawei}. To account for this, we adopt another metric $\frac{\text{\# Correct}}{\text{\# Not Refused}}$~\cite{simpleqa}, which represents the percentage of correctly answered questions out of those that were not refused.
Moreover, we use the relative degradation (RD) of performance (see Eq.~\ref{eq: relative_degradation}) to evaluate a model's visual module.

\begin{table*}[h]
    \small
   \centering
   \renewcommand\arraystretch{1}
   \begin{threeparttable}
    \newcolumntype{?}{!{\vrule width 0.8pt}}
    \setlength{\tabcolsep}{0.8mm}
    \begin{tabular}{l?cccc?cccc?c}
        \toprule
        &\multicolumn{4}{c?}{ \textbf{Text-only QA} }&\multicolumn{4}{c?}{\textbf{Multimodal QA}}&\\
        \cmidrule{2-10}        
        & {Correct}& {Incorrect} & {Refusal}& {$\frac{\text{\# Correct}}{\text{\# Not Refused}}$} & {Correct}& {Incorrect} & {Refusal}& \makecell{$\frac{\text{\# Correct}}{\text{\# Not Refused}}$} & RD\\
        \midrule
        GPT-4o & 77.6 & 22.4 & 0.0 & 77.6 & 67.6 & 29.4 & 3.0 & 69.7 & 12.9\\
        Claude-3.5-Sonnet-20241022 & 77.4 & 15.0 & 7.6 & 83.8 & 61.4 & 26.0 & 12.6 & 70.3 & 20.7\\
        Gemini-2.0-flash-exp & 72.8 & 27.0 & 0.2 & 72.9 & 64.0 & 30.8 & 5.2 & 67.5 & 12.1\\
        \midrule
        LLaVA-OneVision-qwen2-7b-ov & 43.2 & 56.8 & 0.0 & 43.2 & 24.8 & 61.0 & 14.2 & 28.9 & 42.6\\
        Molmo-7B-O-0924 & 42.4 & 57.6 & 0.0 & 42.4 & 32.4 & 66.8 & 0.8 & 32.7 & 23.6\\
        Qwen2-VL-7B-Instruct & 51.0 & 43.2 & 5.8 & 54.1 & 32.2 & 64.4 & 3.4 & 33.3 & 36.9\\
        Qwen2.5-VL-7B-Instruct & 48.6 & 50.8 & 0.6 & 48.9 & 38.4 & 54.0 & 7.6 & 41.6 & 21.0\\
        InternVL2.5-8B & 37.2 & 53.6 & 9.2 & 41.0 & 26.8 & 66.4 & 6.8 & 28.8 & 28.0\\
        Llama-3.2-11B-Vision-Instruct & 50.0 & 24.0 & 26.0 & 67.6 & 40.8 & 45.4 & 13.8 & 47.3 & 18.4\\
        Pixtral-12B-2409 & 53.2 & 44.6 & 2.2 & 54.4 & 34.6 & 63.4 & 2.0 & 35.3 & 35.0\\
        \midrule
        Molmo-72B-0924 & 64.4 & 35.6 & 0.0 & 64.4 & 46.6 & 52.0 & 1.4 & 47.3 & 27.6\\
        Qwen2-VL-72B-Instruct & 62.0 & 31.4 & 6.6 & 66.4 &  37.0 & 59.6 & 3.4 & 38.3 & 40.3\\
        Qwen2.5-VL-72B-Instruct & 63.4 & 30.8 & 5.8 & 67.3 & 51.2 & 42.8 & 6.0 & 54.5 & 19.2\\
        QVQ-72B-Preview & 42.2 & 13.0 & 44.8 & 76.4 & 46.2 & 53.6 & 0.2 & 46.3 & --\\
        InternVL2.5-78B & 52.4 & 40.8 & 6.8 & 56.2 & 37.4 & 50.8 & 11.8 & 42.4 & 28.6\\
        \bottomrule
    \end{tabular}
    \small
    \begin{tablenotes}
       \item Note:
       RD is an abbreviation for the relative degradation of an LVLM's performance when transitioning from text-only QA to multimodal QA. Here the performance is measured by correctness. Notably, QVQ-72B-Preview tends to refuse to respond to text-only questions, which may result in an underestimation of its actual factuality in text-only QA. Therefore, we exclude QVQ-72B-Preview from the RD metric calculation.
    \end{tablenotes}
    \end{threeparttable}
     \caption{Evaluation of various models on \sbench (\%).}\label{tab:exp_overall_evaluation}
\end{table*}

\begin{figure}
	\centering
		\includegraphics[width=1.49in]{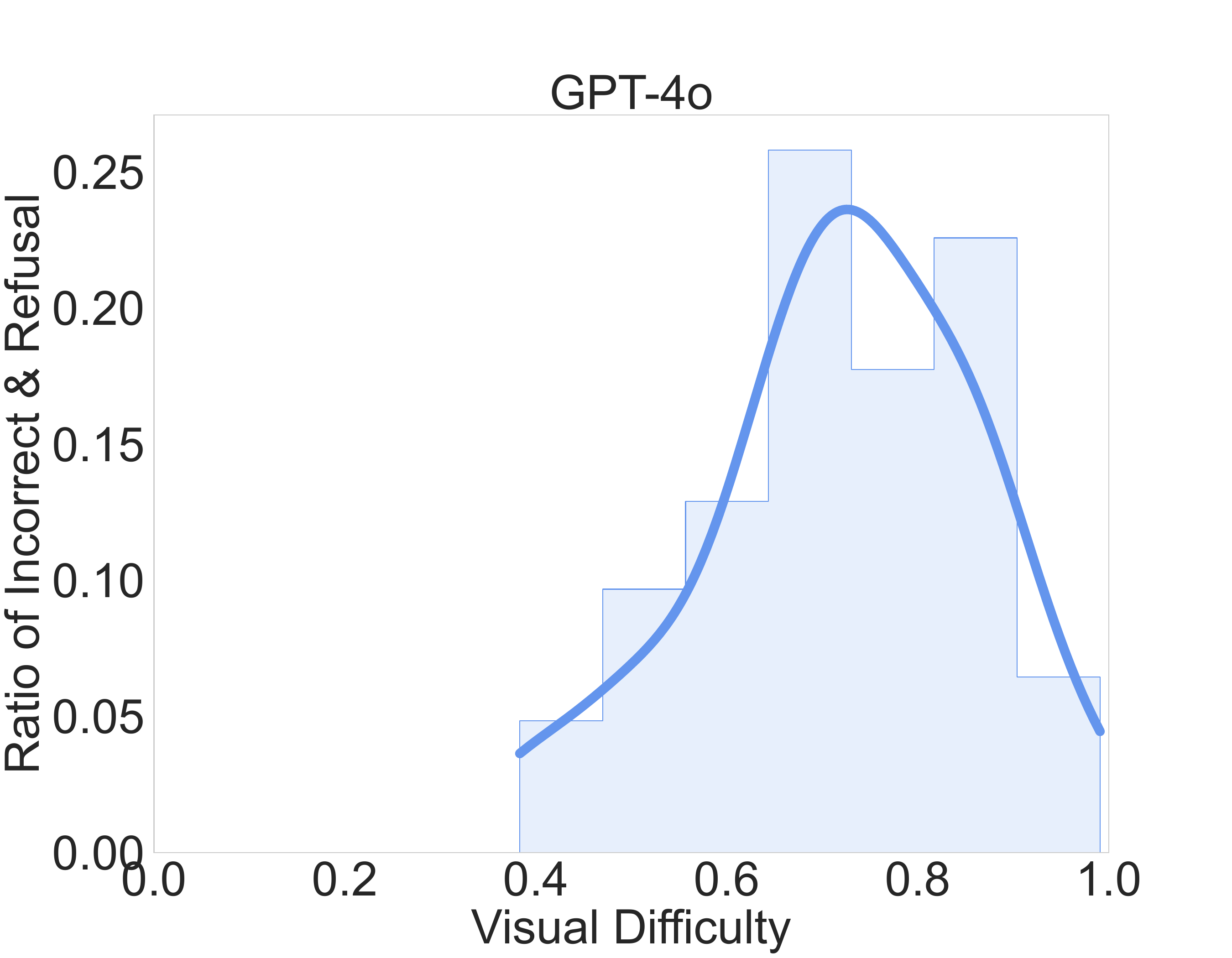}
        \includegraphics[width=1.49in]{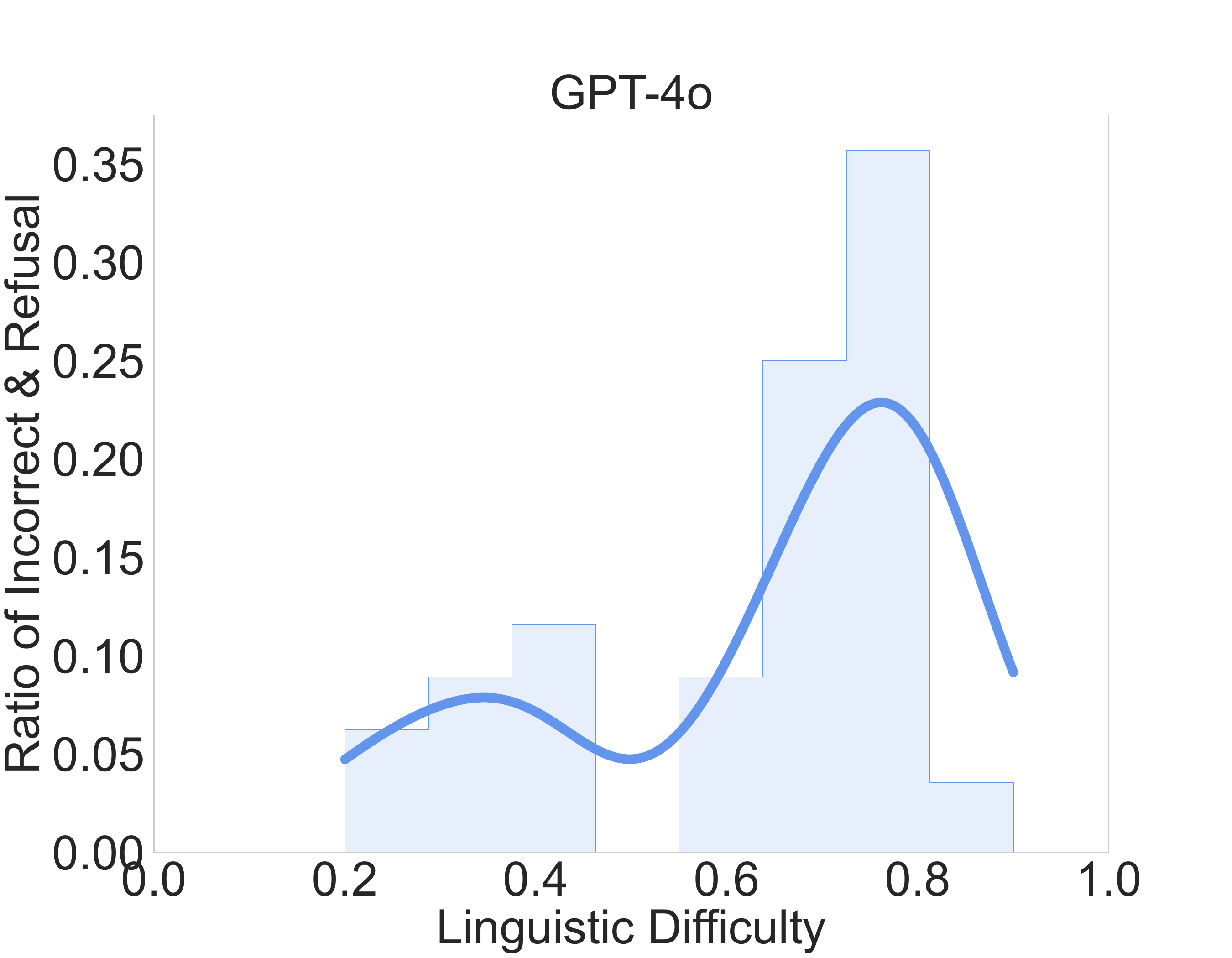}
      \caption{Ratio of failures (incorrect responses and refusals) by GPT-4o across varying difficulty levels. The left sub-figure is based on samples where GPT-4o correctly answers the text-only questions but fails to answer the multimodal questions. The right is based on samples where GPT-4o fails to answer the text-only questions.}
  \label{fig: ratio of incorrectly answered or refused samples by GPT-4o}
\end{figure}

\begin{figure}
	\centering
        \includegraphics[width=1.49in]{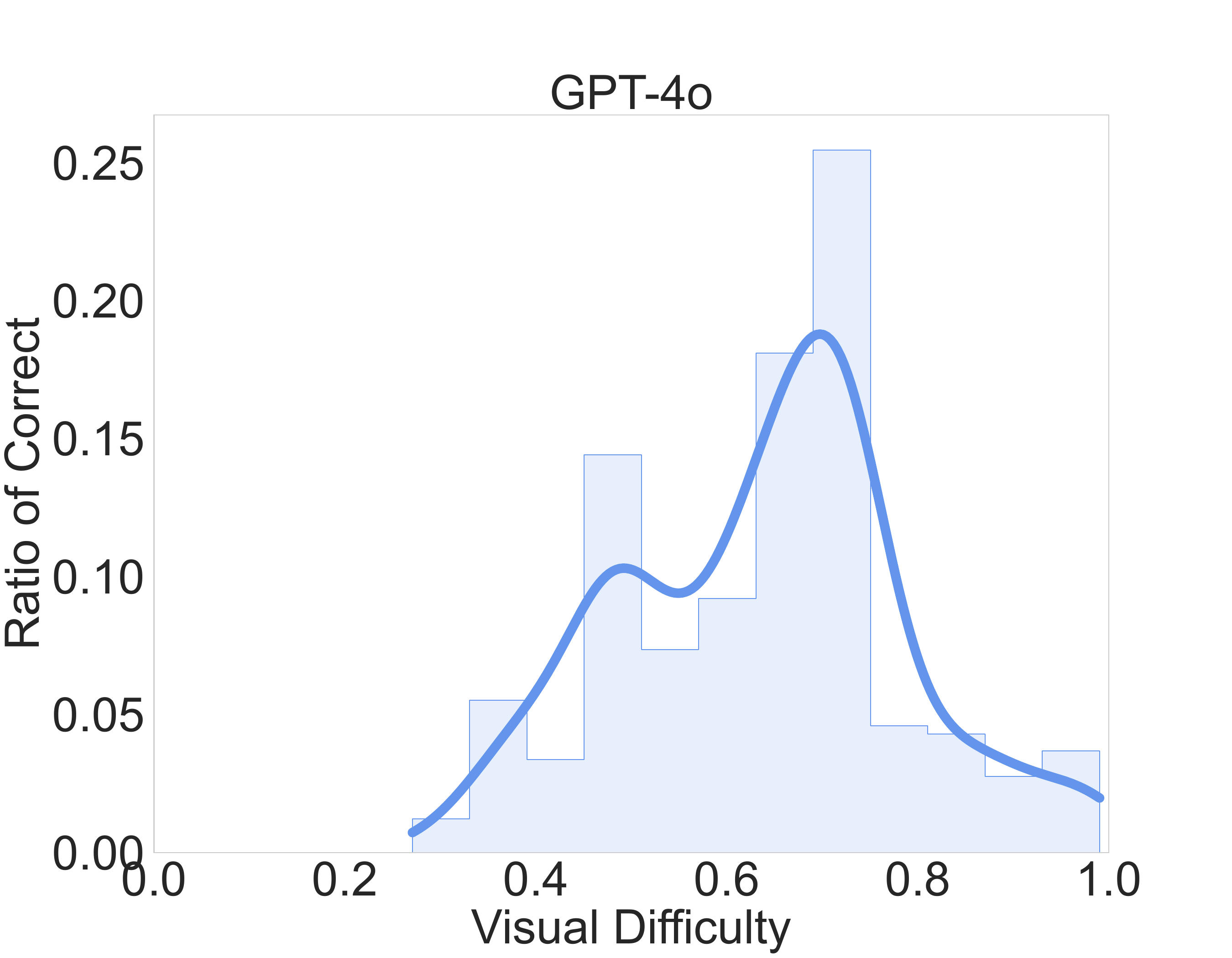}
        \includegraphics[width=1.49in]{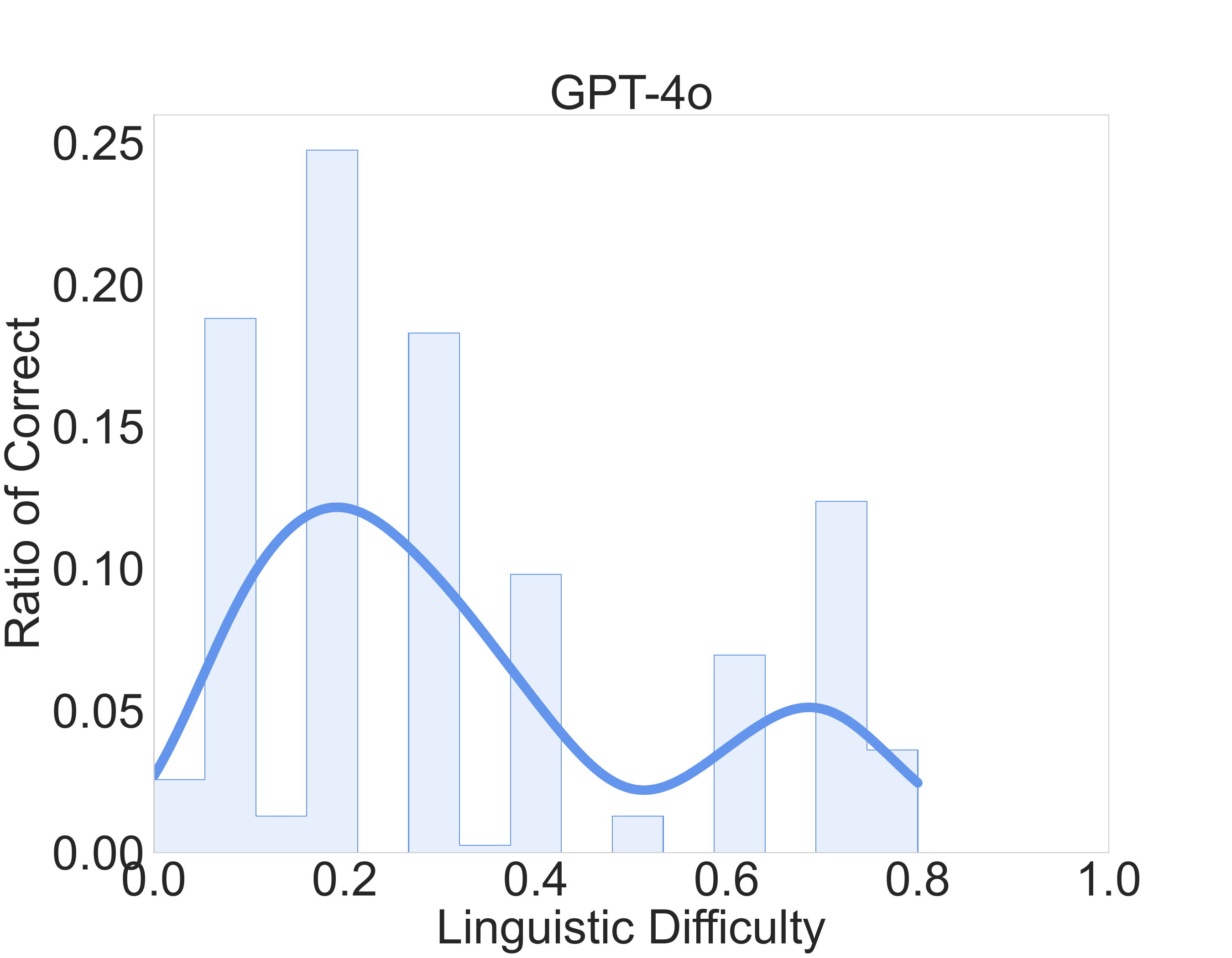}
      \caption{Ratio of correctly answered questions by GPT-4o across varying difficulty levels. The left sub-figure is based on samples where GPT-4o correctly answers the text-only questions and the multimodal questions. The right is based on samples where GPT-4o correctly answer the text-only questions.}
  \label{fig: ratio of success samples by GPT-4o}
\end{figure}

\begin{table*}[h]
    \small
   \centering
   \renewcommand\arraystretch{1}
   \begin{threeparttable}
    \newcolumntype{?}{!{\vrule width 0.8pt}}
    \setlength{\tabcolsep}{0.8mm}
    \begin{tabular}{l?cccc?cccc?c}
        \toprule
        &\multicolumn{4}{c?}{\textbf{Text-only QA }}&\multicolumn{4}{c?}{\textbf{Multimodal QA}}&\\
        \cmidrule{2-10}        
        & {Correct}& {Incorrect} & {Refusal}& {$\frac{\text{\# Correct}}{\text{\# Not Refused}}$} & {Correct}& {Incorrect} & {Refusal}& {$\frac{\text{\# Correct}}{\text{\# Not Refused}}$} & RD\\
        \midrule
        GPT-4o & 56.6 & 43.4 & 0.0 & 56.6 & 37.2 & 55.0 & 7.8 & 40.3 & 34.3\\
        Claude-3.5-Sonnet-20241022 & 54.3 & 28.7 & 17.0 & 65.4 & 37.2 & 38.0 & 24.8 & 49.5 & 31.5\\
        Gemini-2.0-flash-exp & 47.3 & 52.7 & 0.0 & 47.3 & 38.0 & 53.5 & 8.5 & 41.5 & 19.7\\
        \midrule
        Llama-3.2-11B-Vision-Instruct & 24.0 & 24.8 & 51.2 & 49.2 & 20.2 & 52.7 & 27.1 & 27.7 & 15.8\\
        Pixtral-12B-2409 & 31.8 & 64.3 & 3.9 & 33.1 & 17.8 & 78.3 & 3.9 & 18.5 & 44.0\\
        \midrule
        Molmo-72B-0924 & 42.6 & 57.4 & 0.0 & 42.6 & 22.5 & 76.7 & 0.8 & 22.7 & 47.2\\
        Qwen2.5-VL-72B-Instruct & 38.8 & 50.4 & 10.8 & 43.5 & 27.9 & 56.6 & 15.5 & 33.0 & 28.1\\
        \bottomrule
    \end{tabular}
    \end{threeparttable}
        \caption{Evaluation of a part of models on \bench-hard (\%). See Table~\ref{tab:supplementary_exp_hard_overall_evaluation} in Appendix~\ref{appendix: Supplementary experiments} for more results.}\label{tab:exp_hard_overall_evaluation}
\end{table*}

\subsection{Evaluation Results}
\vpara{Overall Evaluation.}
We input multimodal questions and their corresponding text-only questions into LVLMs, respectively. The performance of different models is shown in Table~\ref{tab:exp_overall_evaluation}. We make the following observations:
\textbf{(1) \sbench is a benchmark with challenging samples, capable of differentiating the factuality of different LVLMs.} Even advanced models like GPT-4o achieve only 60\%+ correctness in multimodal QA. Furthermore, substantial performance differences exist across different LVLMs on VisualSimpleQA, which demonstrates the effectiveness of our benchmark in differentiating the capabilities of different LVLMs in multimodal fact-seeking QA.
\textbf{(2) Present frontier LVLMs still require significant improvement on difficult visual recognition tasks.} 
By conducting decoupled evaluation, we observe that these models, particularly open-source models, still have large relative performance degradations when transitioning from text-only QA to multimodal QA.
\textbf{(3) Closed-source frontier LVLMs have clear advantages over current open-source LVLMs.} For instance, GPT-4o outperforms the best-performing open-source model, Qwen2.5-VL-72B-Instruct, by 10.3\% in text-only QA and 15.2\% in multimodal QA in terms of $\frac{\text{\# Correct}}{\text{\# Not Refused}}$, and GPT-4o exhibits a smaller relative performance degradation.
\textbf{(4) Larger models significantly outperform smaller models.} We evaluate different versions of Molmo-0924, Qwen2-VL-Instruction, Qwen2.5-VL-Instruction, and InternVL2.5 with varying parameter sizes. The larger models consistently show better performance in both multimodal and text-only QA tasks.

\vpara{Evaluation on \bench-hard.} 
Table~\ref{tab:exp_hard_overall_evaluation} presents the evaluation on \bench-hard.
\bench-hard increases the difficulty in both visual recognition and knowledge identification.
Concretely, compared to their performance on \bench, each model shows a significant performance drop in text-only QA, highlighting the increased difficulty in knowledge identification. Moreover, larger relative degradations are observed compared to evaluations on \bench, indicating the increased difficulty in visual recognition.

\vpara{Validation of the Difficulty Criteria.} Taking GPT-4o as an example, we analyze the reasonableness of the proposed difficulty criteria.
Based on Eq.~\ref{eq: overall_difficulty}, we further define the visual difficulty as $\text{Avg}\left(1 - R_{\text{norm}}, 1 - PR, \mathbbm{1}_{[\text{fine}]},  \mathbbm{1}_{[\text{w/o}\, \text{TI}]}\right)$, and define the linguistic difficulty as $1 - KP$.
Comparing the results shown in Figure~\ref{fig: ratio of incorrectly answered or refused samples by GPT-4o} and Figure~\ref{fig: ratio of success samples by GPT-4o}, we observe that the failures tend to feature higher visual difficulty.
Similarly, the failures also tend to feature higher linguistic difficulty compared to the correctly answered ones.
Similar observations are observed in other LVLMs (see Figure~\ref{fig: supplementary_ratio of incorrectly answered or refused samples} and Figure~\ref{fig: supplementary_ratio of success samples} in Appendix~\ref{appendix: Supplementary experiments}).

\section{Conclusion}
We introduce \bench, a multimodal fact-seeking QA benchmark for decoupled evaluation of modality-specific modules in LVLMs. Moreover, it includes a series of difficulty criteria to guide human annotation, quantify sample difficulty, and help extract a set of challenging samples, \bench-hard. Experiments with 15 frontier LVLMs demonstrate the effectiveness of \bench. We hope this benchmark can advance research on the factuality of LVLMs.

\section*{Limitations}
This work primarily focuses on fact-seeking questions, which typically feature concise ground truth answers. However, factuality-related challenges also emerge in other contexts, such as visual description, multimodal content creation, and multimodal reasoning. These tasks often involve long-form and non-unique ground truth answers, introducing additional complexities for both benchmark development and model evaluation. We leave the exploration of these aspects to future work.

\bibliography{custom}

\appendix
\onecolumn 
\section{Prompts}\label{sec:appendix_prompts}
\subsection{Prompt for Knowledge Popularity Quantification}\label{appendix: KNOWLEDGE_POPULARITY_PROMPT}
\begin{tcolorbox}[title={\parbox{\textwidth}{\raggedright \textbf{\textcolor{green}{KNOWLEDGE\_POPULARITY\_PROMPT}}}}]

\texttt{KNOWLEDGE\_POPULARITY\_PROMPT} = """ \\
    You are a knowledgeable, honest, and impartial AI assistant.\\
    
    Here is the question: \\
\texttt{\{text\_only\_question\}} \\

Here is the ground truth answer to the question: \\
\texttt{\{ground\_truth\_answer\}} \\

Based on the extensive knowledge you have acquired, please assess the popularity of the knowledge required to answer this question in the training corpora. Provide a popularity score between 0.0 and 1.0, with a higher score indicating higher popularity.\\

Only output the popularity Score: \\
    """
\end{tcolorbox}

\subsection{Prompts for Automatic Evaluation by DeepSeek-V3}\label{appendix: Prompts for Automatic Evaluation by DeepSeek-V3}
\begin{tcolorbox}[title={\parbox{\textwidth}{\raggedright \textbf{\textcolor{green}{SYSTEM\_PROMPT\_FOR\_EVALUATION}}}}]

\texttt{SYSTEM\_PROMPT\_FOR\_EVALUATION} = """ \\
    You are a helpful AI assistant. Given a question, a ground truth answer (Answer\_1), and an answer predicted by a large model (Answer\_2), your task is to assess the correctness of Answer\_2 based on the following criteria:\\

1. Label Answer\_2 as 1 (correct) if it contains all the information conveyed by Answer\_1 and does not include any incorrect information. If the wording of Answer\_2 differs from Answer\_1 but conveys the same information, Answer\_2 should still be considered correct.\\
2. Label Answer\_2 as 0 (refusal) if it explicitly states that the model cannot answer the question.\\
3. Label Answer\_2 as -1 (incorrect) if it contains incorrect information or omits any information conveyed by Answer\_1, and it does not explicitly state that the model cannot answer the question.\\

Note: Incorrect information refers to any details in response to the question that contradict or diverge from the information provided in Answer\_1.

Only output the label: 1, 0, or -1.\\
    """
\end{tcolorbox}

\begin{tcolorbox}[title={\parbox{\textwidth}{\raggedright \textbf{\textcolor{green}{PROMPT\_FOR\_EVALUATION}}}}]

\texttt{PROMPT\_FOR\_EVALUATION} = """ \\
    Question: \texttt{\{text\_only\_question\}}\\
    Answer\_1: \texttt{\{ground\_truth\_answer\}}\\
    Answer\_2: \texttt{\{response\}}\\
    Label: \\
    """
\end{tcolorbox}
\section{Display of Examples}\label{sec:appendix_examples}
\begin{figure*}
    \centering
    \includegraphics[width=6.in]{figures/examples_1.pdf}
\end{figure*}

\begin{figure*}
    \centering
    \includegraphics[width=6.in]{figures/examples_2.pdf}
\end{figure*}

\begin{figure*}
    \centering
    \includegraphics[width=6.in]{figures/examples_3.pdf}
\end{figure*}

\clearpage
\section{Modality-Specific Modules in LVLMs}
\label{appendix:different modules in LVLMs}

\begin{table}[h]
    \small
   \centering
   \renewcommand\arraystretch{1}
   \begin{threeparttable}
    \newcolumntype{?}{!{\vrule width 0.8pt}}
    \setlength{\tabcolsep}{2.5mm}
    \begin{tabular}{l?cc}
        \toprule
        & \textbf{\makecell{Linguistic Module}} & \textbf{\makecell{Visual Module}}\\
        \midrule
        GPT-4o & unknown & unknown \\
        \midrule
        Claude-3.5-Sonnet-20241022 & unknown & unknown \\
        \midrule
        Gemini-2.0-flash-exp & unknown & unknown \\
        \midrule
        LLaVA-OneVision-qwen2-7b-ov & Qwen2-7B & SigLIP-SO400M \\
        \midrule
        Molmo-7B-O-0924 & OLMo-7B-1024-preview & OpenAI’s ViT-L/14 336px \\
        \midrule
        Qwen2-VL-7B-Instruct & Qwen2-7B & Qwen2-ViT-675M \\
        \midrule
        InternVL2.5-8B & InternLM2.5-7B-Chat & InternViT-300M-448px-V2.5 \\
        \midrule
        Qwen2.5-VL-7B-Instruct & Qwen2.5-7B & Qwen2.5-ViT w/ window attention \\
        \midrule
        Llama-3.2-11B-Vision-Instruct &  Llama-3.1-8B & ViT-H/14 \\
        \midrule
        Pixtral-12B-2409 & Mistral-NeMo-12B & Pixtral-ViT-400M \\
        \midrule
        Molmo-72B-0924 & Qwen2-72B & OpenAI’s ViT-L/14 336px \\
        \midrule
        Qwen2-VL-72B-Instruct & Qwen2-72B & Qwen2-ViT-675M \\
        \midrule
        QVQ-72B-Preview & Qwen2-72B & ViT-675M \\
        \midrule
        InternVL2.5-78B & Qwen2.5-72B-Instruct & InternViT-6B-448px-V2.5 \\
        \midrule
        Qwen2.5-VL-72B-Instruct & Qwen2.5-72B & Qwen2.5-ViT w/ window attention \\
        \bottomrule
    \end{tabular}
    \end{threeparttable}
    \caption{Modality-specific modules in different LVLMs.}
    \label{tab: different modules in LVLMs}
\end{table}
\section{Supplementary Experimental Results}\label{appendix: Supplementary experiments}

\begin{table*}[h]
    \small
   \centering
   \renewcommand\arraystretch{1}
   \begin{threeparttable}
    \newcolumntype{?}{!{\vrule width 0.8pt}}
    \setlength{\tabcolsep}{1.1mm}
    \begin{tabular}{l?cccc?cccc?c}
        \toprule
        &\multicolumn{4}{c?}{\textbf{Text-only QA }}&\multicolumn{4}{c?}{\textbf{Multimodal QA}}&\\
        \cmidrule{2-10}        
        & {Correct}& {Incorrect} & {Refusal}& {$\frac{\text{\# Correct}}{\text{\# Not Refused}}$} & {Correct}& {Incorrect} & {Refusal}& {$\frac{\text{\# Correct}}{\text{\# Not Refused}}$} & \makecell{RD}\\
        \midrule
        LLaVA-OneVision-qwen2-7b-ov & 18.6 & 81.4 & 0.0 & 18.6 & 8.5 & 65.1 & 26.4 & 11.6 & 54.3\\
        Molmo-7B-O-0924 & 17.1 & 82.9 & 0.0 & 17.1 & 11.6 & 87.6 & 0.8 & 11.7 & 32.2\\
        Qwen2-VL-7B-Instruct & 24.0 & 64.3 & 11.7 & 27.2 & 14.7 & 80.6 & 4.7 & 15.4 & 38.8\\
        Qwen2.5-VL-7B-Instruct & 23.3 & 75.2 & 1.5 & 23.6 & 17.1 & 62.0 & 20.9 & 21.6 & 26.6\\
        InternVL2.5-8B & 17.8 & 62.8 & 19.4 & 22.1 & 8.5 & 74.4 & 17.1 & 10.3 & 52.2\\
        \midrule
        Qwen2-VL-72B-Instruct & 39.5 & 48.1 & 12.4 & 45.1 & 21.7 & 73.6 & 4.7 & 22.8 & 45.1\\
        QVQ-72B-Preview & 16.3 & 19.4 & 64.3 & 45.7 & 20.2 & 79.1 & 0.7 & 20.3 & --\\
        InternVL2.5-78B & 24.8 & 59.7 & 15.5 & 29.4 & 14.0 & 56.6 & 29.4 & 19.8 & 43.5\\
        \bottomrule
    \end{tabular}
    \end{threeparttable}
        \caption{Evaluation of models on \bench-hard (\%).}\label{tab:supplementary_exp_hard_overall_evaluation}
\end{table*}

\begin{figure*}[h]
	\centering
		\includegraphics[width=2in]{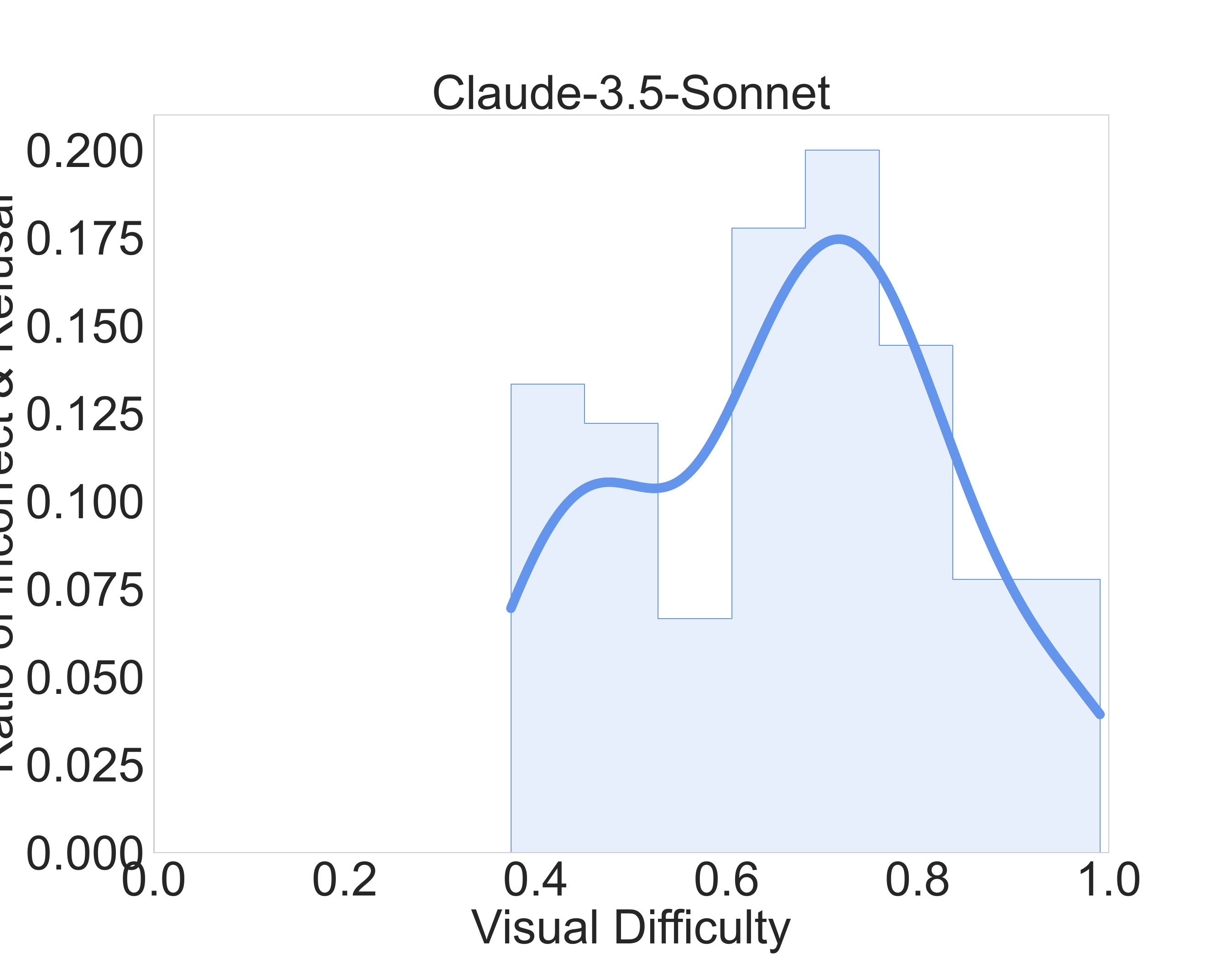}
         \includegraphics[width=2in]{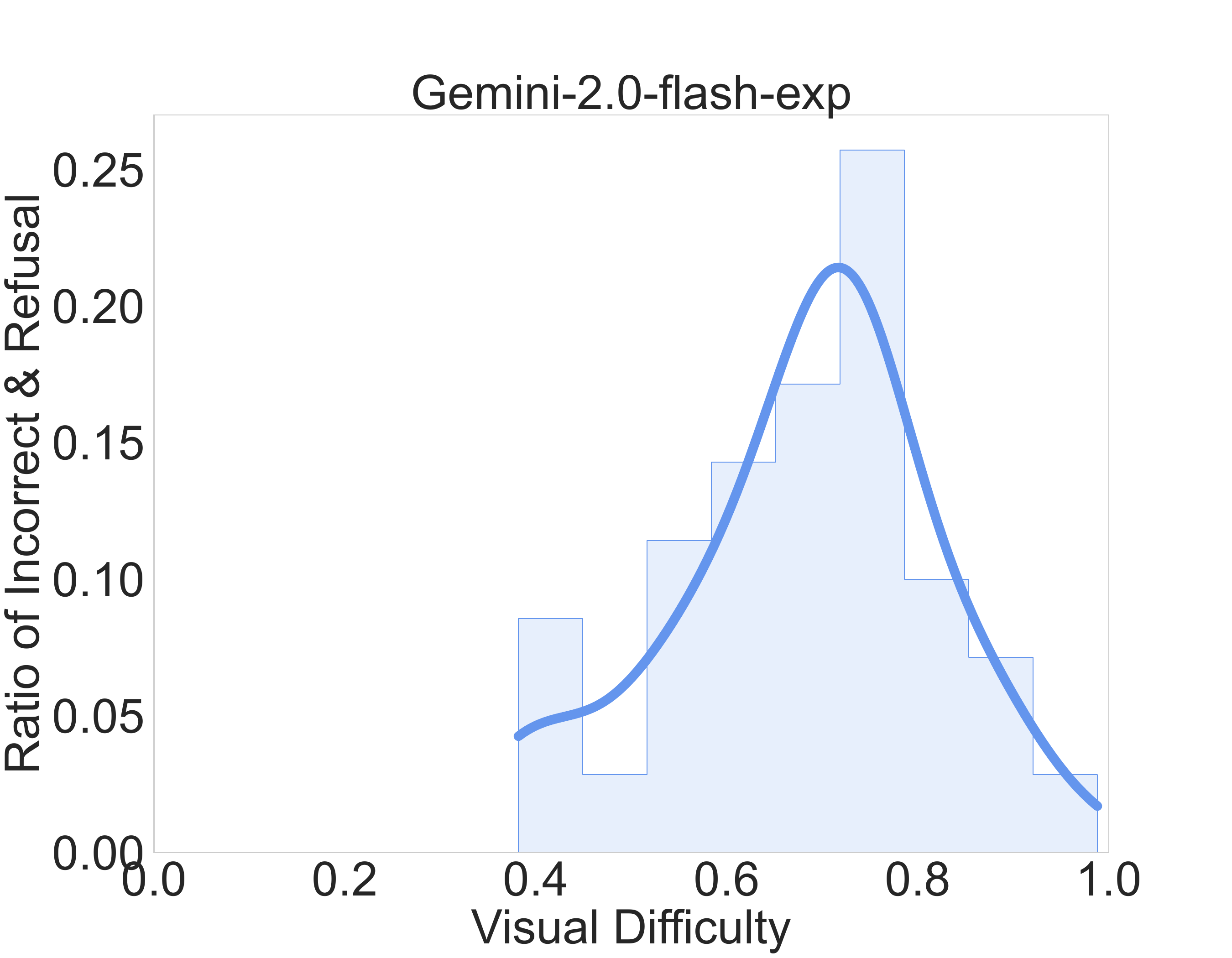}
        \includegraphics[width=2in]{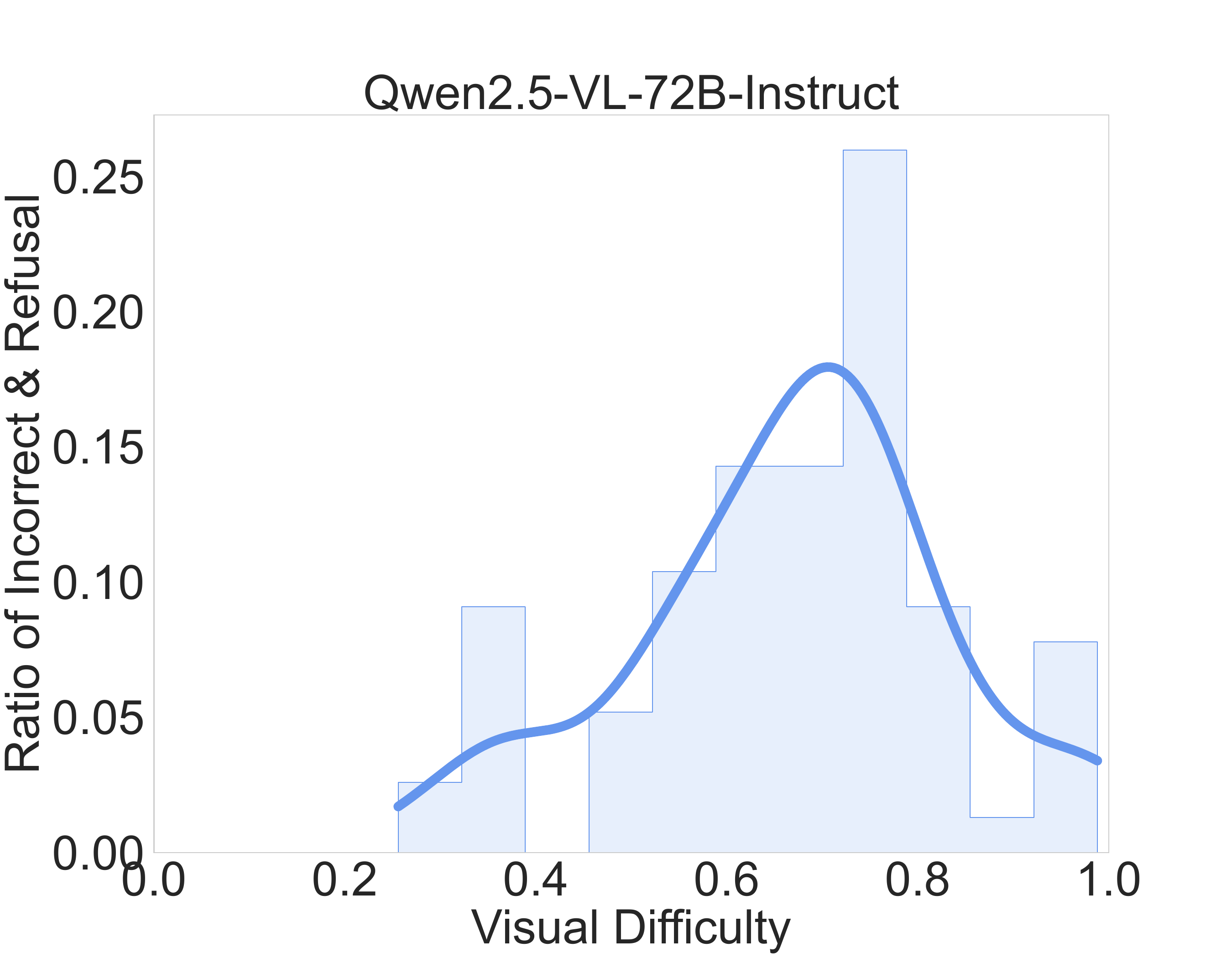}
        
        \includegraphics[width=2in]{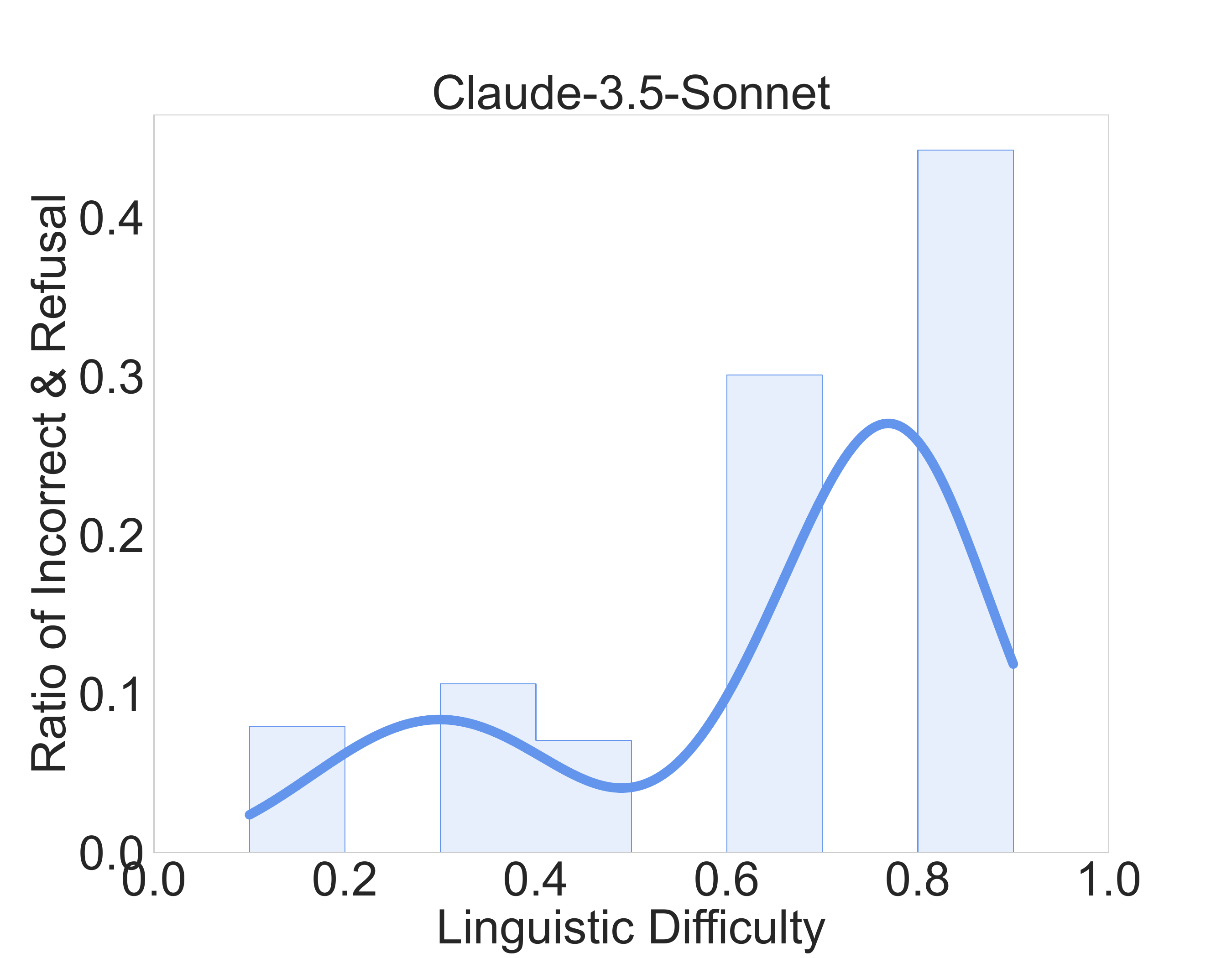}
        \includegraphics[width=2in]{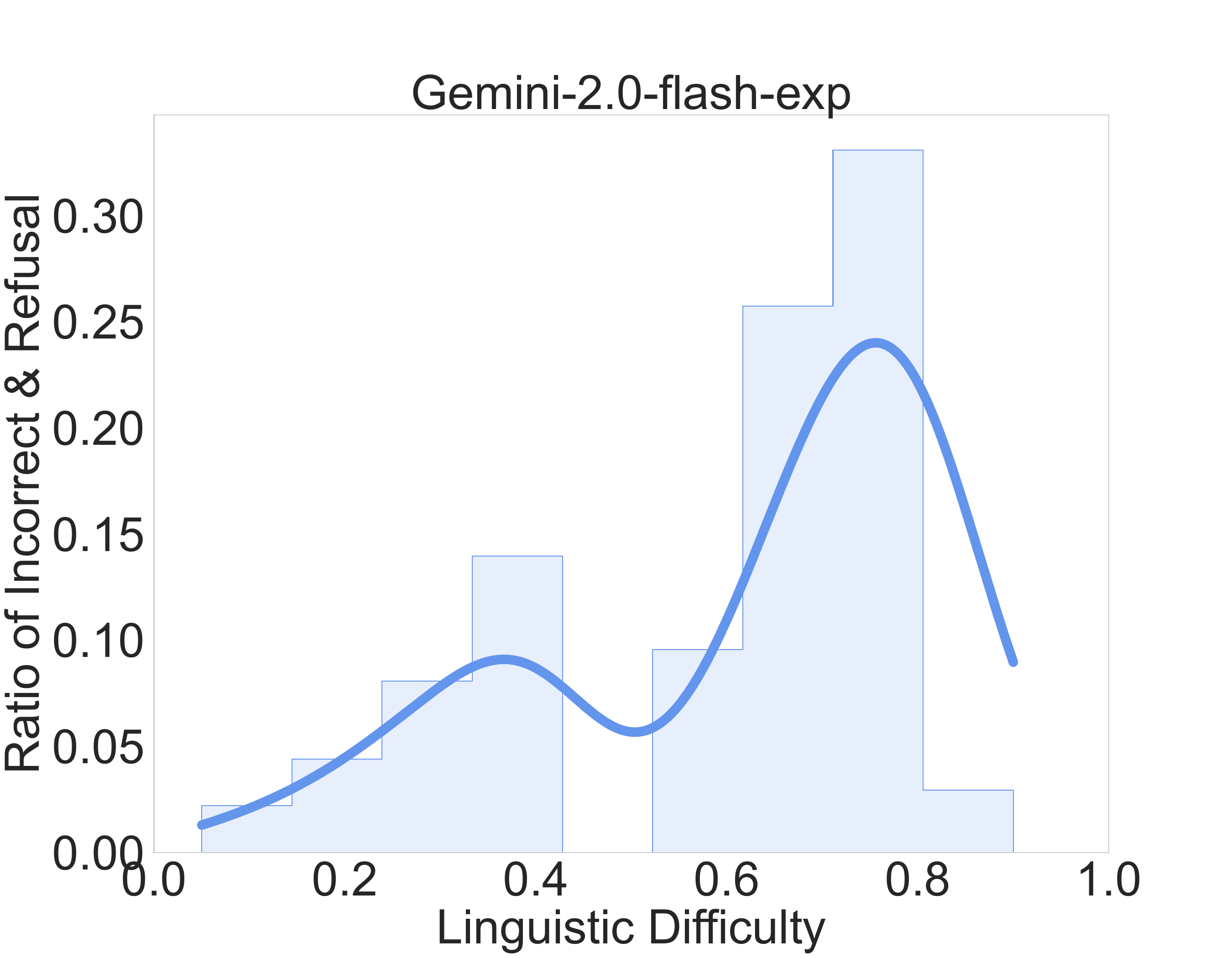}
        \includegraphics[width=2in]{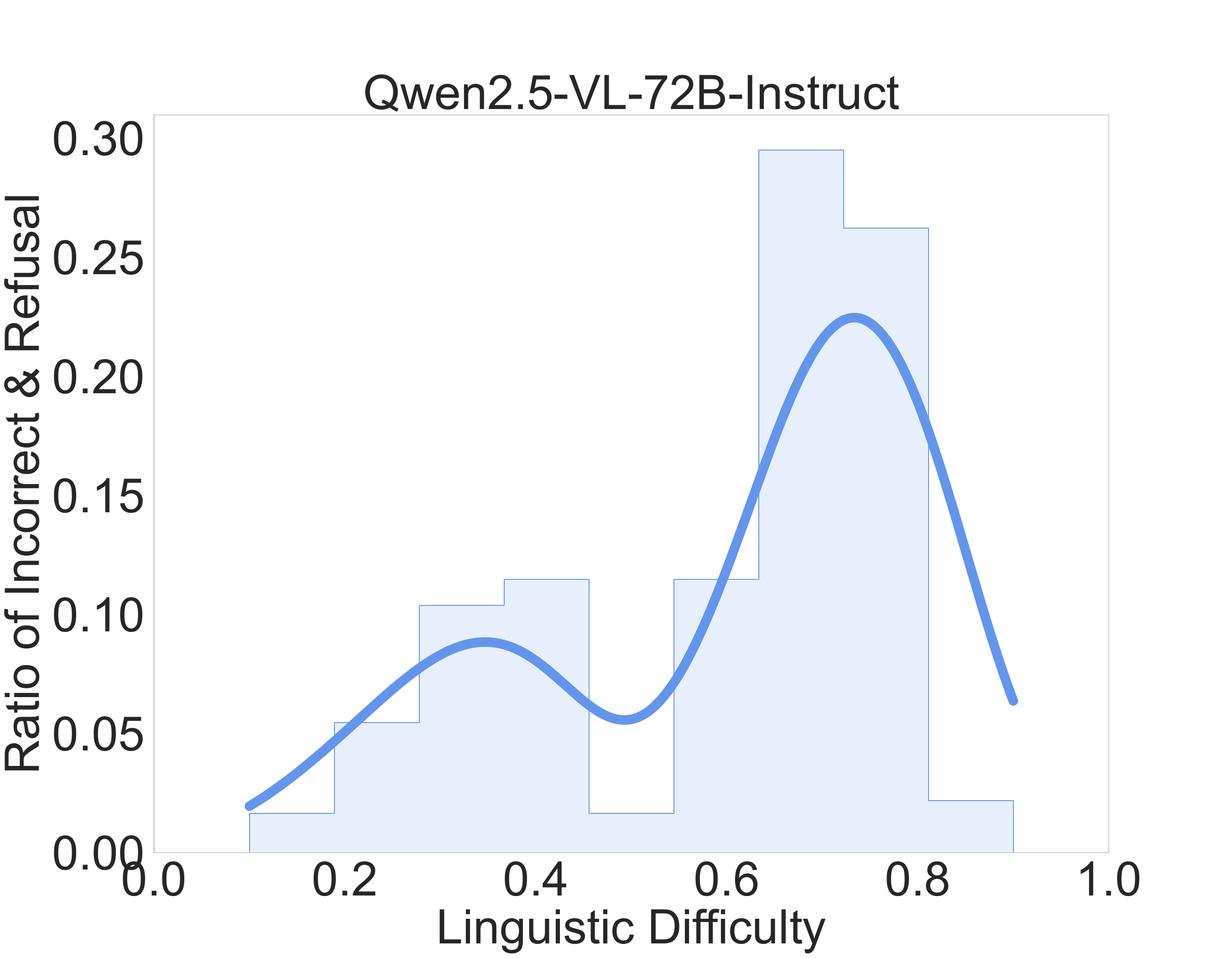}
      \caption{Ratio of failures (incorrect responses and refusals) across varying difficulty levels. The upper sub-figures are based on samples where models correctly answer the text-only questions but fail to answer the multimodal questions. The bottom sub-figures are based on samples where models fail to answer the text-only questions.}
      \label{fig: supplementary_ratio of incorrectly answered or refused samples}
\end{figure*}

\begin{figure*}[h]
	\centering
        \includegraphics[width=2in]{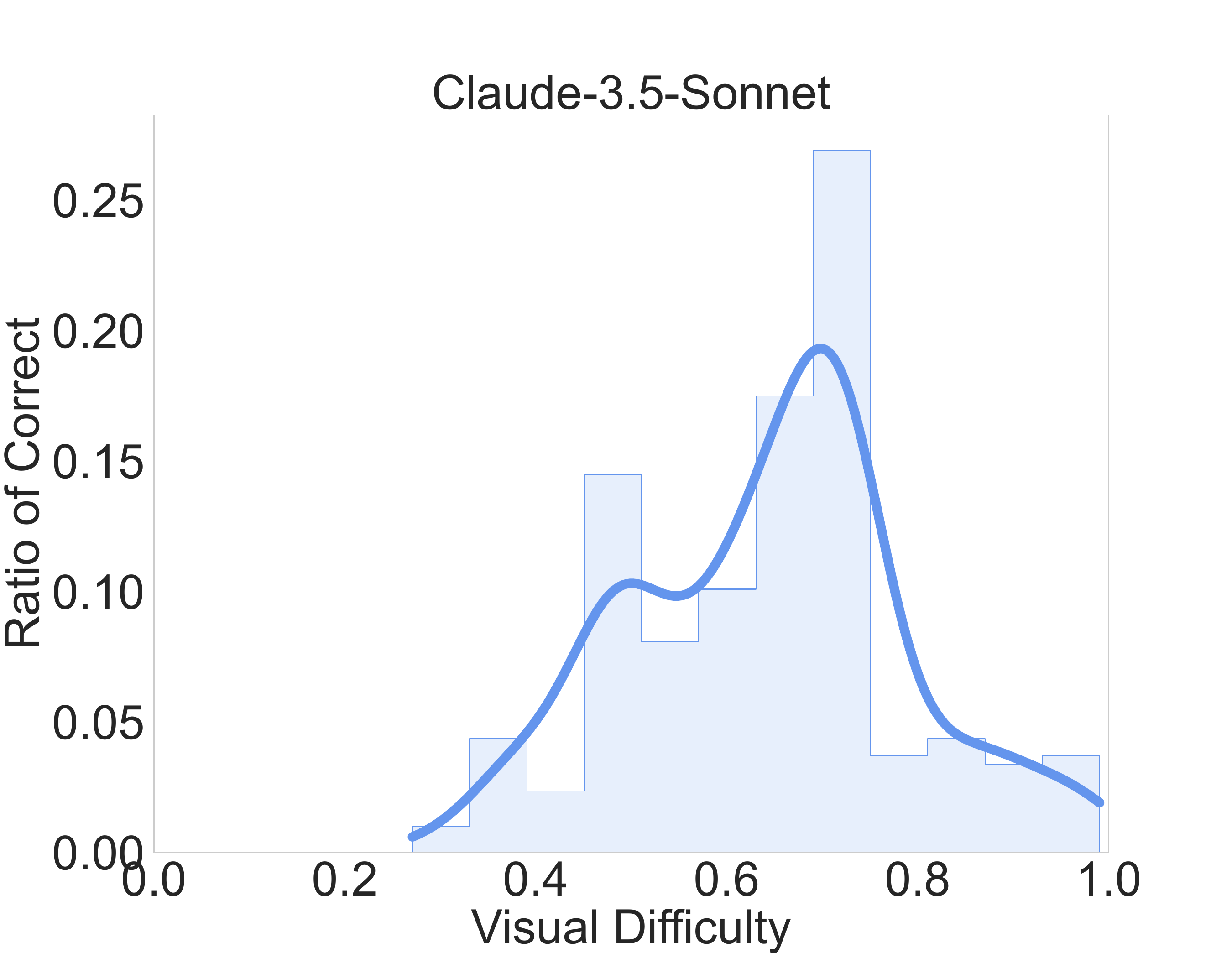}
        \includegraphics[width=2in]{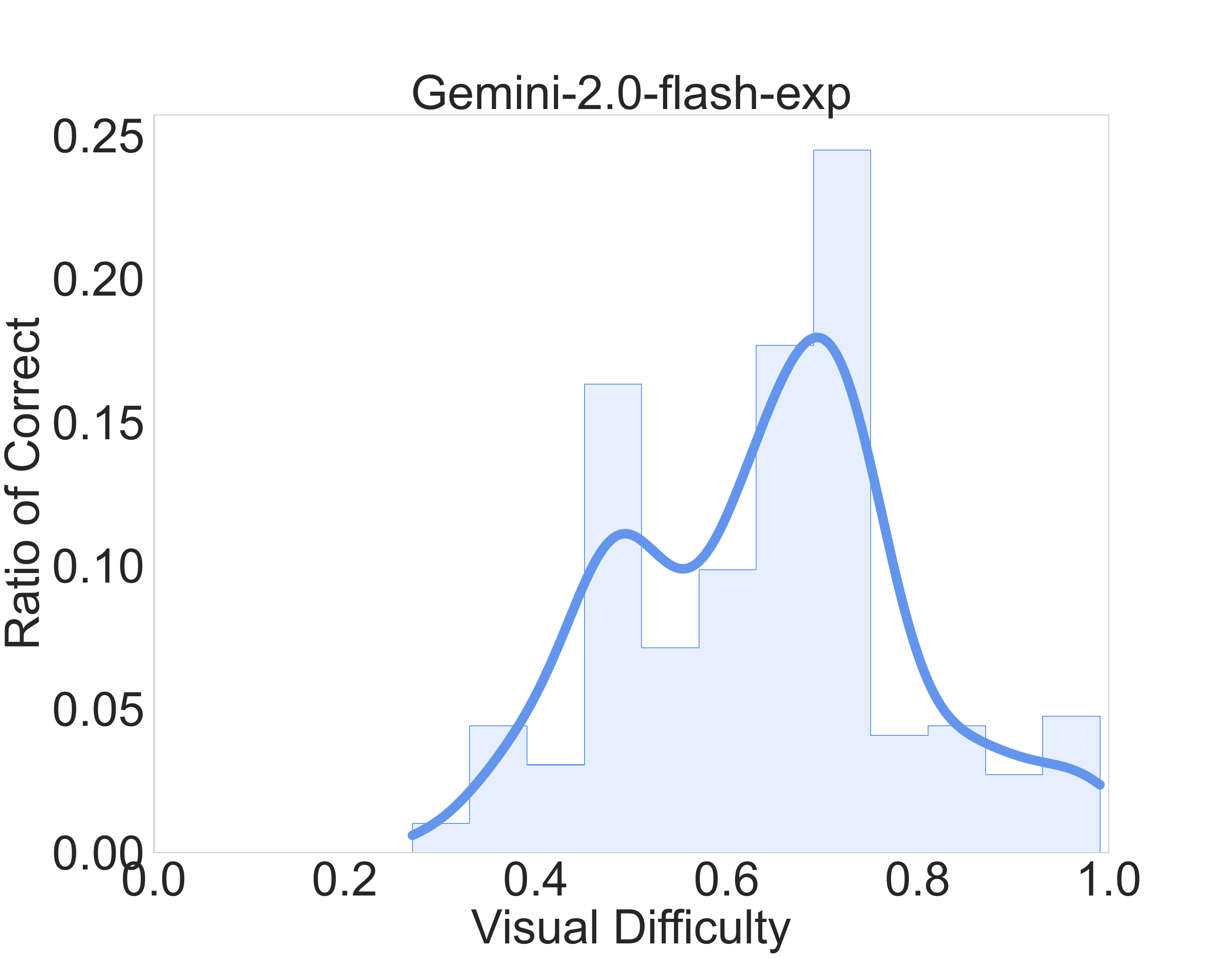}
        \includegraphics[width=2in]{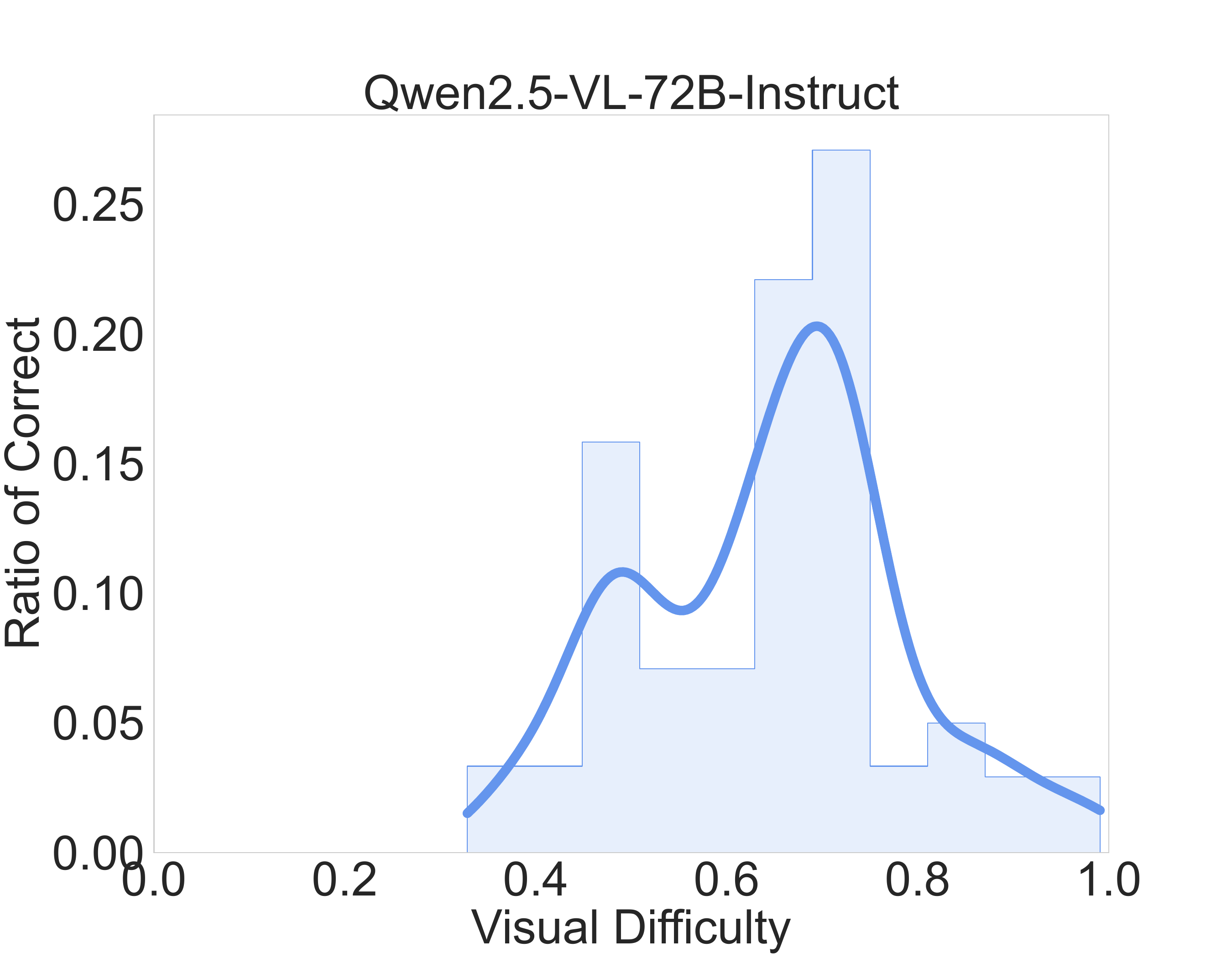}
        
        \includegraphics[width=2in]{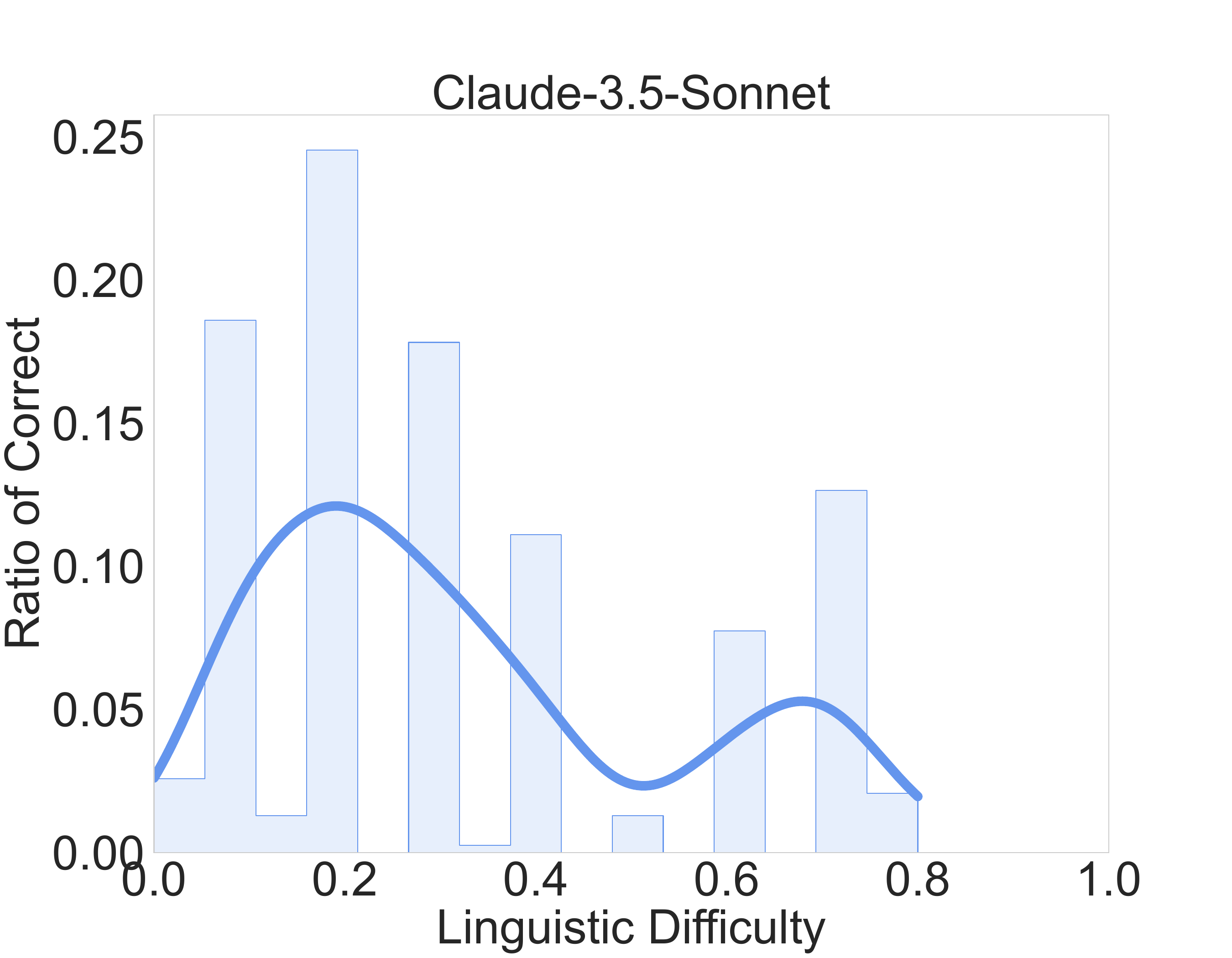}
        \includegraphics[width=2in]{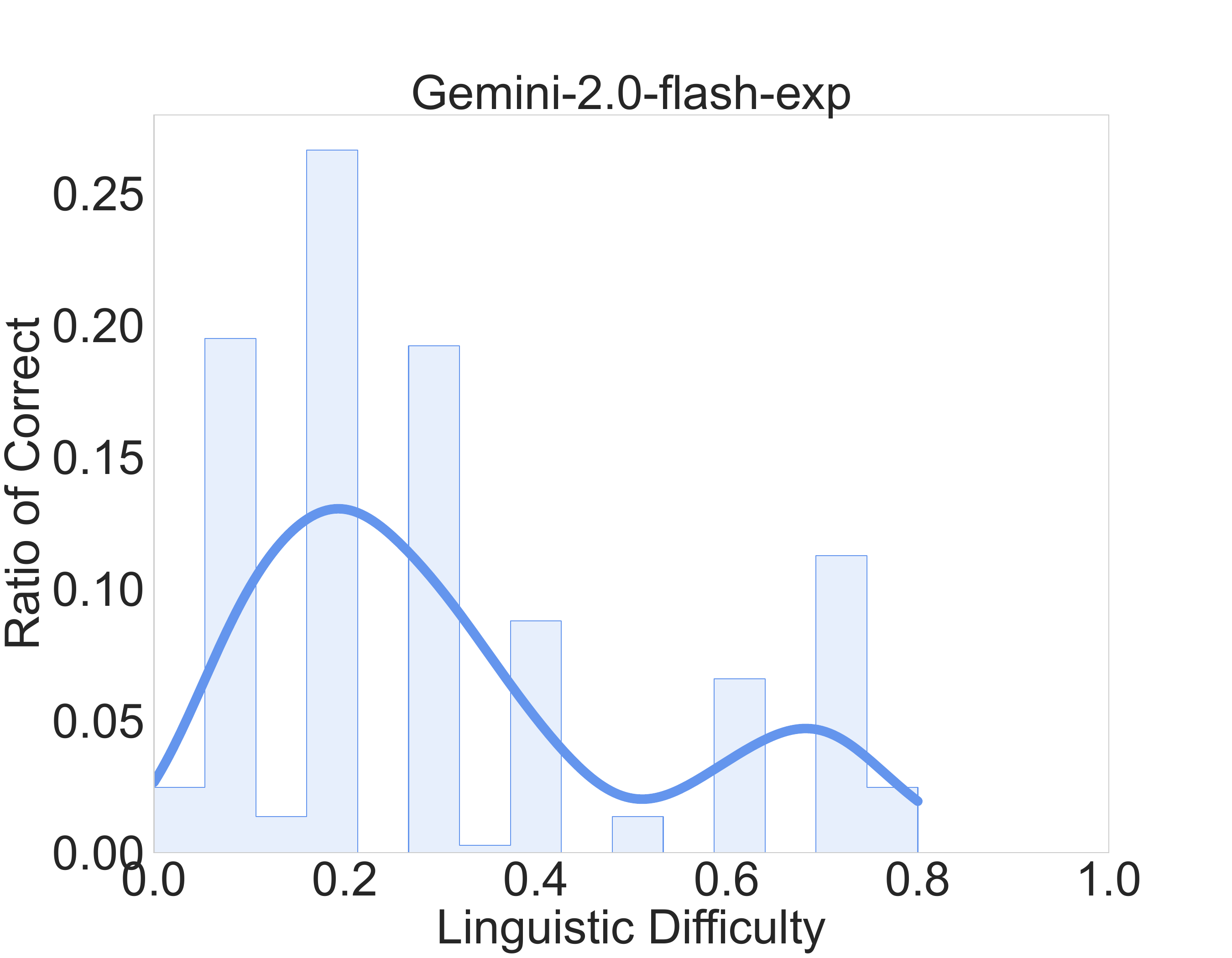}
        \includegraphics[width=2in]{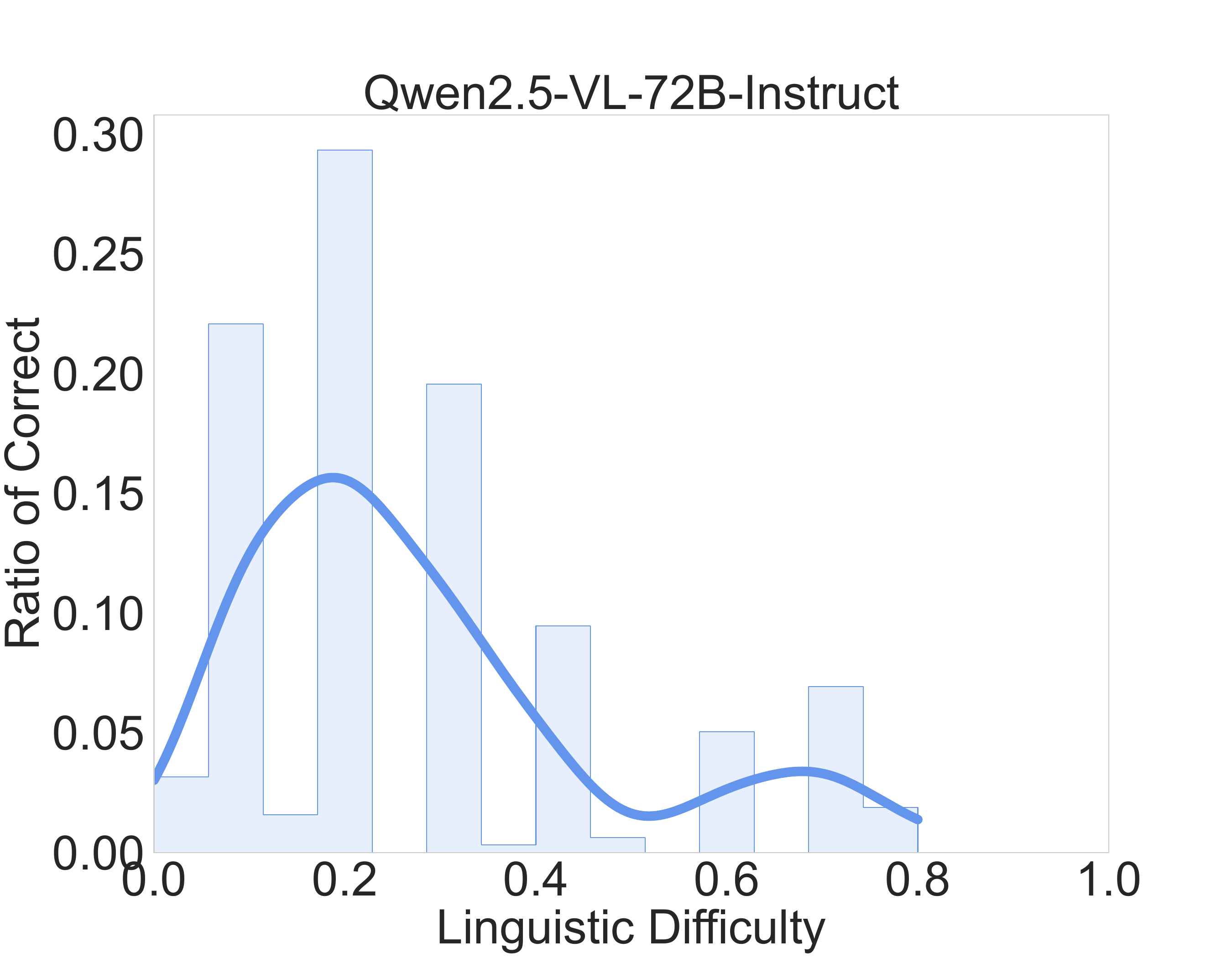}
        
      \caption{Ratio of correctly answered questions across varying difficulty levels. The upper sub-figures are based on samples where models correctly answer the text-only questions and the multimodal questions. The bottom sub-figures are based on samples where models correctly answer the text-only questions.}
      \label{fig: supplementary_ratio of success samples}
\end{figure*}

\end{document}